\pdfoutput=1

\documentclass[11pt]{article}
\usepackage{subcaption}

\usepackage{acl}

\usepackage{times}
\usepackage{latexsym}

\usepackage[T1]{fontenc}
\usepackage[utf8]{inputenc}
\usepackage{microtype}
\usepackage{inconsolata}

\usepackage{xcolor}
\usepackage{colortbl}
\usepackage{graphicx}
\usepackage{caption}
\usepackage{subcaption}
\usepackage[most]{tcolorbox}
\usepackage{cuted}
\usepackage{booktabs}
\usepackage{amsmath}
\usepackage{amssymb}
\usepackage{framed}

\newcommand{\modelname}[1]{\texttt{#1}}

\title{Coreference as an indicator of context scope in multimodal narrative}

\author{
  Nikolai Ilinykh\textsuperscript{\textdagger},\,
  Shalom Lappin\textsuperscript{\textsection \textdagger},\,
  Asad Sayeed\textsuperscript{\textdagger},
  \and Sharid Lo\'{a}iciga\textsuperscript{\textdagger} \\
  \textsuperscript{\textdagger}Dept. of Philosophy, Linguistics, and Theory of Science, University of Gothenburg \\
  \textsuperscript{\textsection}School of Electronic Engineering and Computer Science, Queen Mary University of London \\
  \textsuperscript{\textsection}Dept. of Informatics, King's College London \\
  \texttt{nikolai.ilinykh@gu.se, s.lappin@qmul.ac.uk,} \\
  \texttt{asad.sayeed@gu.se, sharid.loaiciga@gu.se}
}

\begin{document}
\maketitle

\begin{abstract}
We demonstrate that large multimodal language models differ substantially from humans in the distribution of coreferential expressions in a visual storytelling task. 
We introduce a number of metrics to quantify the characteristics of coreferential patterns in both human- and machine-written texts. 
Humans distribute coreferential expressions in a way that maintains consistency across texts and images, interleaving references to different entities in a highly varied way.
Machines are less able to track mixed references, despite achieving perceived improvements in generation quality.
Materials, metrics, and code for our study are available at \url{https://github.com/GU-CLASP/coreference-context-scope}.
\end{abstract}

\section{Introduction}

Generative models produce text that many perceive as increasingly human-like.
However, machine-generated text conceals important distinctions to which people are sensitive \citep{russell2025peoplefrequentlyusechatgpt}.
Work on visual narrative shows that there is still a gap between human and machine ability to generate coherent text \citep{xu-etal-2018-skeleton}. 

A key difference between human and machine writing behaviour is the distribution of coreferential elements in English.
We find that texts generated in a multimodal task setting have considerably different distributions of transitions between {\it coreference chains}, i.e., chains of referring expressions pointing to the same entity.
In this work, we describe our methodology, using a set of metrics that we apply to state-of-the-art multimodal language models in a visual storytelling task.
All tested models show substantial differences in measured behaviour with respect to human-generated reference texts.
This points to a need for a quantitative approach to coherence that accounts for the characteristics of coreference in texts generated from images by machines.
A proper handling of reference has implications for the ability of these models to perform multi-modal grounding, reasoning and inference tasks in the way humans expect \citep{MckookRatcliff-1992}.
Our contributions are:

\vspace{-.1cm}
\begin{itemize}
\setlength{\itemsep}{0pt}
\setlength{\parskip}{0pt}
\setlength{\parsep}{0pt}
\item We introduce a set of distributional metrics capturing coreference transition patterns.
\item We evaluate recent multimodal models on visual storytelling with our metrics.
\item We perform an analysis of multimodal alignment of character consistency in text.
\end{itemize}
\vspace{-.15cm}

%It is not by chance that LLMs produce text that is ``human-like''. 
%However, 
Humans build stories in a certain way, focusing on visual events and the logical connections and participants involved in them.
Event participants take the form of characters in the stories, with some reappearing as the story unfolds. %with some being brought forward at different points as the story develops.
This character introduction and reprisal is precisely what coreference resolution encodes.
Coreference resolution identifies referring expressions or \textit{mentions} in a long text and chains them into distinct entities or \textit{coreference chains} \citep{AR-2016}.
The type of mention is also influenced by the salience of a character -- whether they are newly introduced or already known \citep{Prince1992,grosz-etal-1995-centering}.
%Furthermore, the salience of those characters at any given point as new or known affects referring expression type \citep{Prince1992,grosz-etal-1995-centering}.

Work on generation of visual narratives is often evaluated on automatic scores such as BLEU \citep{papineni-etal-2002-bleu} or ROUGE \citep{lin-2004-rouge}, but they correlate poorly with human judgements \citep{hsu-etal-2022-learning}.
Other metrics that look at trigram repetitions \citep{goldfarb-tarrant-etal-2020-content} or differences in distributions \citep{pillutla_2021} have a higher correlation with fluency as perceived by humans.
However, these metrics do not capture the \textit{tellability} of a story.
In this study we use surface-level reference patterns to capture key aspects of tellability, including continuity, salience, and character switching.
Our main contribution is a diagnostic framework for evaluating narrative consistency, demonstrating how coreference-based features can distinguish between human and model-generated stories in a multimodal setting.
%Moreover, our MCC (Multimodal Character Continuity) metric bridges vision and language, going beyond pure text-level analysis. 

\subsection{Related work}

\paragraph{Coherence} A coherent text consists of logically connected utterances, maintained through cohesive elements like lexical similarity and discourse connectives \citep{HallidayHasan1976}.
However, since coherence is not easily ``extractable'', its evaluation often relies on tasks like automatic summarisation \citep{barzilay-lapata-2008-modeling}, sentence perturbation \citep{dini-etal-2025-text}, or sentence intrusion detection \citep{shen-etal-2021-evaluating}.

\paragraph{On relation to Centering Theory}
Our work focuses on describing how character references in visual narratives behave, and how this can reflect coherence and tellability.
While we do not directly build on formal discourse theories, we think that the concepts of information structure and the role of topic and focus together with insights from Centering Theory \citep{grosz-etal-1995-centering} are relevant.
However, we chose not to rely on Centering Theory directly, as it imposes constraints such as assumptions about anaphora resolution \citep{lappin1994} and requires parameter tuning, including concepts like ``utterance'' \citep{poesio2004} which it is non-trivial to define.
As shown by \citet{chai2022}, Centering Theory can complement but not replace modern neural coreference systems.
Since our stories are grounded in image sequences, directly applying Centering Theory poses additional challenges that we hope to explore in future work.

\paragraph{Visual storytelling}
%Visual storytelling recently became a viable task.
Visual storytelling is emerging as a key testbed for evaluating grounded language models.
The point of departure has been series of static images with captions collected either from photo albums \citep{huang-etal-2016-visual} or movie scenes \citep{hong_visual_2023}.
Generating tellable stories requires identifying common elements across images and consistently referring to them.
Previous work has focused on creation of character-centric stories \citep{liu_generating_2024,liu-detecting-2023} and on optimisation of the loss function for coherence without an explicit concept of coreference \citep{hong-etal-2023-visual}.
%Visual storytelling is also important in daily communication (e.g., narrating visual scenes, summarising events, explaining diagrams).
Visual storytelling poses unique challenges and provides a window into coherence, common sense reasoning, and discourse grounding.
Because stories are structured around sequences of events and characters in images, they provide a rich but controlled environment for probing model capabilities -- including reference tracking, focus management, and multimodal alignment.

\paragraph{Coreference} There is a long tradition of text-only systems for coreference resolution \citep{coref-survey-23}.
More recently, reference relationships beyond texts have been explored in simulated environments where agents or participants interact with the environment or each other \citep{lee-etal-2022-learning}.
This type of setting elicits reference relations that are difficult to find in texts, such as changes in perspective and reference meaning negotiation between participants \citep{tang-etal-2024-grounding}.

\section{Methodology}

\begin{table}[ht]
    \centering
    \scriptsize
    \begin{tabular}{l@{\hspace{4pt}}ccc@{\hspace{8pt}}ccc}
        \toprule
        \textbf{Model} 
        & \multicolumn{3}{c}{\textbf{\# sentences}} 
        & \multicolumn{3}{c}{\textbf{\# words}} \\
        \cmidrule(lr){2-4} \cmidrule(lr){5-7} 
        & \(\mu\) & \(\downarrow\) & \(\uparrow\) & \(\mu\) & \(\downarrow\) & \(\uparrow\) \\
        \midrule
        \modelname{DeepSeek-VL2-4.5B} & 25.91 & 8 & 52   & 417.08 & 140 & 522 \\
        \modelname{DeepSeek-VL2-1B}   & 13.26 & 1 & 69   & 265.39 & 1   & 573 \\
        \modelname{Gemini 2.0 Flash}  & 8.18  & 1 & 24   & 104.76 & 3   & 456 \\
        \modelname{GPT4o}             & 12.51 & 6 & 21   & 212.99 & 114 & 321 \\
        \modelname{InternVL2.5-78B}   & 15.88 & 1 & 71   & 302.24 & 1   & 1021 \\
        \modelname{Qwen2-VL-72B}      & 14.33 & 4 & 47   & 230.01 & 60  & 516 \\
        \modelname{Qwen2-VL-7B}       & 11.89 & 4 & 32   & 223.83 & 66  & 511 \\
        \addlinespace[0.1em]
        \hline
        \addlinespace[0.1em]
        Human                         & 6.67  & 4 & 23   & 85.39  & 56  & 332 \\
        \bottomrule
    \end{tabular}
    \caption{
    General descriptive statistics for generated and human outputs: number of sentences and total word counts per output. Columns show mean (\(\mu\)), minimum (\(\downarrow\)), and maximum (\(\uparrow\)) values.}
    \label{tab:generalstats}
\end{table}

We examine how large language models handle coreference by generating text from the same prompts used for human-written stories.
Using a strong automatic coreference resolver, we annotate both model and human outputs and compute our metrics on them.
This setup enables a direct comparison of coreferential behaviour, allowing us to identify differences between models and humans.

%We test the coreferential behaviour of LLMs generating text from the same writing prompts as used for human references.
%We apply a high-performing automatic coreference detector to the model outputs, and compute several metrics for the generated annotations.
%We also use the same detector and metrics for the original human output.
%This allows us to identify substantial differences in behaviour.

\subsection{Generation}
VWP \citep{hong_visual_2023} is a collection of human-written narratives obtained by presenting participants with a curated sequence of up to $10$ images from MovieNet \citep{huang2020movienetholisticdatasetmovie}, yielding sequences of images paired with one human-written story each.
We choose to work with VWP because its stories are written and evaluated by humans to be tellable, diverse, and grounded.
\citet{hong_visual_2023} also show that VWP offers better semantic cohesion and coherence than VIST \citep{huang-etal-2016-visual} for the same image sequences.

We use the image sequences to generate text stories from several multimodal models.
The prompts are provided in Appendix~\ref{app:prompts} and technical details are provided in Appendix~\ref{sec:techdetails}.
We then employ \modelname{Link-Append} \citep{bohnet_coreference_2023} to annotate both the machine- and human-generated texts.
This means that for each sequence of images in our evaluation dataset, we have a story generated by a human, stories generated by multiple LLMs, and lists of coreference chains for all stories.
Table~\ref{tab:generalstats} provides statistics for all the stories.%\footnote{We use grey background in our tables to indicate models that are not open-sourced.}

\subsection{Data}
\label{sec:data}
VWP has a total of $12,627$ examples with pre-defined train/dev/test splits.
To ensure a representative evaluation while limiting computational demands, we performed stratified sampling and selected $30\%$ or $3,786$ examples for evaluation.
The sampling is proportional to the distribution in the original splits and based on two factors: the number of stories per movie, and the number of images per story.
Appendix~\ref{sec:data_distr} provides additional statistics.

\subsection{Models}
\label{sec:models}

We employ two versions of \modelname{DeepSeek-VL2} ($1$B and $4.5$B of activated parameters, \citet{wu_deepseek-vl2_2024}), two versions of \modelname{Qwen2-VL} ($7$B and $72$B parameters, \citet{wang_qwen2-vl_2024}), \modelname{InternVL2.5-78B} \citep{chen_expanding_2025}, \modelname{Gemini 2.0 Flash}\footnote{\url{https://deepmind.google/technologies/gemini/}}, and \modelname{GPT4o}\footnote{\url{https://openai.com/index/hello-gpt-4o/}} (version \texttt{gpt-4o-2024-08-06}).
We also use the decoder-based \modelname{Link-Append} system \citep{bohnet_coreference_2023} for automatic coreference resolution.
This system is based on the $13$B-parameter \modelname{mT5} \citep{xue-etal-2021-mt5} and processes only text.
\modelname{Link-Append} has a reported performance of $83.3$ CoNLL score for English. 
While discriminative models may achieve slightly better scores \citep{martinelli2024}, we chose Link-Append for its compatibility with our task, where end-to-end generation aligns naturally with sequence-based modelling.

\begin{table*}[ht]
    \centering
    \begin{tabular}{l@{\hspace{3pt}}ccc@{\hspace{6pt}}ccc@{\hspace{6pt}}ccc@{\hspace{6pt}}ccc}
        \toprule
        \textbf{Model} 
        & \multicolumn{3}{c}{\textbf{\# words-as-mentions}} 
        & \multicolumn{3}{c}{\textbf{\# chains}} 
        & \multicolumn{3}{c}{\textbf{chain size}} 
        & \multicolumn{3}{c}{\textbf{CCI}} \\
        \cmidrule(lr){2-4} \cmidrule(lr){5-7} \cmidrule(lr){8-10} \cmidrule(lr){11-13}
        & \(\mu\) & \(\downarrow\) & \(\uparrow\) & \(\mu\) & \(\downarrow\) & \(\uparrow\) & \(\mu\) & \(\downarrow\) & \(\uparrow\) & \(\mu\) & \(\downarrow\) & \(\uparrow\) \\
        \midrule
        \modelname{DeepSeek-VL2-4.5B}   & 135.60 & 35 & 323   & 10.75 & 2 & 26   & 13.06 & 6   & 97.5 & 2.58 & 0 & 15.92 \\
        \modelname{DeepSeek-VL2-1B}     & 77.17  & 0  & 410   & 6.98  & 0 & 28   & 11.33 & 0   & 333  & 1.62 & 0 & 18.30 \\
        \modelname{Gemini 2.0 Flash}    & 27.48  & 0  & 147   & 3.88  & 0 & 16   & 6.47  & 0   & 27   & 0.83 & 0 & 6.60 \\
        \modelname{GPT4o}               & 52.84  & 11 & 131   & 6.16  & 1 & 15   & 8.89  & 3.2 & 27   & 1.62 & 0 & 7.64 \\
        \modelname{InternVL2.5-78B}     & 87.85  & 0  & 294   & 7.65  & 0 & 22   & 11.53 & 0   & 33.5 & 1.79 & 0 & 10.20 \\
        \modelname{Qwen2-VL-72B}        & 66.84  & 7  & 199   & 6.69  & 1 & 17   & 10.23 & 2.2 & 36   & 1.47 & 0 & 7.57 \\
        \modelname{Qwen2-VL-7B}         & 76.64  & 5  & 311   & 6.62  & 1 & 19   & 11.71 & 2.5 & 35.4 & 1.80 & 0 & 11.20 \\
        \addlinespace[0.1em]
        \hline
        \addlinespace[0.1em]
        Human                           & 26.09  & 2  & 176   & 3.85  & 1 & 14   & 6.87  & 2   & 23.5 & 0.89 & 0 & 6.43 \\
        \bottomrule
    \end{tabular}
    \caption{Descriptive statistics across models: number of words identified as mentions by LinkAppend, number of coreference chains, average chain size, and Chain Crossing Index (CCI). Columns show mean (\(\mu\)), minimum (\(\downarrow\)), and maximum (\(\uparrow\)) values.}
    \label{tab:generalstats2}
\end{table*}

\begin{table*}[ht]
    \centering
    %\scriptsize
    \begin{tabular}{
        l@{\hspace{6pt}}c@{\hspace{6pt}}c@{\hspace{6pt}}c@{\hspace{6pt}}c@{\hspace{6pt}}c
        @{\hspace{6pt}}c@{\hspace{6pt}}c
        @{\hspace{6pt}}c@{\hspace{6pt}}r@{}l
    }
        \toprule
        \textbf{Model} 
        & \textbf{CharTr} & \textbf{CharDr} & \textbf{CharAd} & \textbf{CharCh} & \textbf{CharRe} 
        & \multicolumn{2}{c}{\textbf{MCC}} 
        & \textbf{REC} & \multicolumn{2}{c}{\boldmath\(\rho\)} \\
        \cmidrule(lr){7-8} \cmidrule(lr){10-11}
        & & & & & & \(\mu\) & \(\downarrow\) & & \multicolumn{2}{c}{} \\
        \midrule
        \modelname{DeepSeek-VL2-4.5B} & 0.06 & 0.90 & 0.88 & 0.57 & 0.57 & 0.76\textsuperscript{\dag} & 0.20 & 0.61 & -0.045 & \textbf{**} \\
        \modelname{DeepSeek-VL2-1B}   & 0.03 & 0.91 & 0.88 & 0.46 & 0.33 & 0.72\textsuperscript{\dag} & 0.16 & 0.61 & -0.045 & \textbf{**} \\
        \modelname{Gemini 2.0 Flash}  & 0.10 & 0.84 & 0.82 & 0.48 & 0.48 & 0.78\textsuperscript{\dag} & 0.11 & 0.57 &  0.352 & \textbf{**} \\
        \modelname{GPT4o}             & 0.10 & 0.85 & 0.83 & 0.54 & 0.60 & 0.76\textsuperscript{\dag} & 0.21 & 0.65 & -0.009 &             \\
        \modelname{InternVL2.5-78B}   & 0.13 & 0.80 & 0.78 & 0.42 & 0.58 & 0.74\textsuperscript{\dag} & 0.18 & 0.64 &  0.012 &             \\
        \modelname{Qwen2-VL-72B}      & 0.12 & 0.82 & 0.80 & 0.45 & 0.63 & 0.79\textsuperscript{\dag} & 0.29 & 0.64 &  0.113 & \textbf{**} \\
        \modelname{Qwen2-VL-7B}       & 0.15 & 0.77 & 0.74 & 0.37 & 0.61 & 0.78\textsuperscript{\dag} & 0.22 & 0.57 &  0.131 & \textbf{**} \\
        \addlinespace[0.1em]
        \hline
        \addlinespace[0.1em]
        Human                         & 0.23 & 0.67 & 0.63 & 0.27 & 0.54 & \hspace{-.15cm}0.84 & 0.29 & 0.65 &  0.005 &             \\
        \bottomrule
    \end{tabular}
    \caption{Aggregated qualitative metric values and referring expression change (REC) across models.
    MCC is shown with mean (\(\mu\)) and minimum (\(\downarrow\)). Pearson correlation (\(\rho\)) shows correlation between REC and text length. Values marked with \textbf{**} denote statistical significance (\(p < 0.05\)). 
    Mean values marked with \textsuperscript{\dag} differ significantly from human according to a two-sample t-test.}
    \label{tab:char_cont}
\end{table*}

\subsection{Quantitative metrics}
\label{sec:quant}

We identify character coreference chains by string-matching VWP character names with \modelname{Link-Append} mentions, labelling the entire chain as a character chain if at least one match is found.
Our metrics are computed on the sentence level.

\paragraph{Character transition (CharTr)}  
Let \( C_s \) be the set of character coreference chains in sentence \( s \). For each consecutive sentence pair \( (s, s+1) \), we define the indicator \( T_s \) as \( T_s = 1 \) if \( C_s \cap C_{s+1} \neq \emptyset \), and \( T_s = 0 \) otherwise.
A higher value indicates that consecutive sentences tend to share at least one character, implying character continuity.

\paragraph{Character drop (CharDr)}
For each sentence pair, if \( C_s \) is non-empty, the drop ratio is defined as \( \text{CharDr} = \frac{|C_s \setminus C_{s+1}|}{|C_s|} \).
This metric represents the proportion of characters that disappear from one sentence to the next, with higher values indicating less character continuity.

\paragraph{Character addition (CharAd)}
If \( C_{s+1} \) is non-empty, the addition ratio is \( \text{CharAd} = \frac{|C_{s+1} \setminus C_s|}{|C_{s+1}|} \).
A higher value indicates that many new characters are introduced in the next sentence.

\paragraph{Character change (CharCh)}
For pairs where both \( C_s \) and \( C_{s+1} \) are non-empty, we define \( \text{CharCh}_s = 1 \) if \( C_s \cap C_{s+1} = \emptyset \), and 0 otherwise.
The metric captures the proportion of sentence pairs with a complete change of characters.

\paragraph{Character reappearance (CharRe)}
For each character chain \( c \), \( s_{\min}(c) \) and \( s_{\max}(c) \) are the first and last sentences in which it appears.
Normalised by the maximum possible spread (\( N-1 \)), the reappearance metric is \( \text{CharRe} = \frac{1}{|\mathcal{C}|} \sum_{c \,\in\, \mathcal{C}} \frac{s_{\max}(c) - s_{\min}(c)}{N-1} \).
Higher values mean characters reappear in distant sentences.

\paragraph{Multimodal character continuity}
We introduce a metric to quantify how consistently each character is referenced across text and images.  
For each character \( C \), we compute text continuity \( T_C \) as the fraction of sentences between the first (\( s_{\min} \)) and last (\( s_{\max} \)) mention of \( C \) that actually include \( C \) using coreference chains.  
Similarly, image continuity \( I_C \) is the fraction of images between the first (\( j_{\min} \)) and last (\( j_{\max} \)) appearance of \( C \) that include \( C \) based on bounding box annotations.  
While \( T_C \) and \( I_C \) are modality-specific, they are aligned over the same sequence length: each story has the same number of text descriptions and images.  
This alignment enables direct comparison, as both metrics are normalised over equivalent spans.  
We define continuity consistency as \( 1 - |T_C - I_C| \), reflecting the agreement between modalities for each character.  
The final multimodal character continuity (MCC) score for a story is the average continuity across all characters.  
We compute MCC for every story and compare distributions across sources (e.g., human vs.\ model) using two-sample \( t \)-tests and Cohen’s \( d \) for effect size.  
Details on visual character detection are provided in Appendix~\ref{app:mmc_details}.

\paragraph{Referring expression change (REC)}
The \texttt{REC} metric captures how consistently a character chain is realised across mentions (e.g., as a proper name or pronoun).  
For a character chain \( c \), let the mention sequence be \( \text{MS}(c) = [m_1, m_2, \dots, m_k] \), where each \( m_i \) is a proper name (N), pronoun (P), or both.  
We set \(\texttt{REC}(c) = 0\) if all mentions are realised the same way (\( |\{\text{MS}(c)\}| = 1 \)), and \(\texttt{REC}(c) = 1\) if the form changes at least once.  
Higher \texttt{REC} values indicate more variation in referring expressions, while lower values suggest consistent usage.  
We also compute Pearson correlation between \texttt{REC} and text length (word count).

%The referring expression change (\texttt{REC}) measures how a character coreference chain is realised across mentions (e.g., as a proper name or a pronoun).
%For a character chain \( c \), let its mention sequence be \( \text{MS}(c) = [m_1, m_2, \dots, m_k] \), where each \( m_i \) is a proper name (N), pronoun (P), or both.
%We define \(\texttt{REC}(c) = 0\) if all mentions are realised the same way (\( |\{\text{MS}(c)\}| = 1 \)), and \(\texttt{REC}(c) = 1\) if the realisation changes at least once.  
%A higher \(\texttt{REC}\) indicates greater variation in referring expressions, while a lower value suggests more consistent usage.
%We also compute the Pearson correlation between \(\texttt{REC}\) and text length (total word count).

\section{Results and analysis}
\label{sec:results}

We begin by examining the results in Table~\ref{tab:generalstats2}, with example outputs provided in Appendices~\ref{sec:examples} -- \ref{sec:examples3}.
Humans refer to fewer entities on average ($3.85$), but each is mentioned multiple times ($6.87$) suggesting a focused narrative.
The chain crossing index (CCI) measures how often two chains intersect, excluding overlaps or disjoint chains. 
A low human CCI ($0.89$) reflects consistent reference to key characters, with less frequent switching between entities.

\modelname{Gemini 2.0 Flash} is closest to humans across all metrics, though it sometimes produces $0$ mentions, indicating inconsistency in referencing characters.
While its structure appears human-like, it may omit key entities entirely.
In contrast, \modelname{DeepSeek-VL2-4.5B} generates more entities, longer chains, and more frequent cross-chain intersections (high CCI), reflecting over-generation.
\modelname{GPT4o} produces fewer chains overall but tracks characters more consistently than \modelname{DeepSeek-VL2-4.5B}.

The results in Table~\ref{tab:char_cont} show that \textit{human}-generated stories show stronger character continuity across sentences.
Humans have higher transition scores (e.g., $0.23$), drop and add characters less frequently (e.g., $0.67$ and $0.63$), and rarely switch to entirely new sets of characters between sentences.
Characters are also reintroduced after shorter gaps (mean $0.54$) unlike in many model outputs.
These patterns suggest that human narratives maintain a more coherent and trackable set of characters, while models tend to drop, add, or switch characters more often.

This trend is reinforced by MCC scores, where humans outperform all models, indicating stronger alignment between text and image character mentions.
Two-sample t-tests confirm the difference is significant in every case (\(p < 0.05\)), with effect sizes (Cohen’s \(d\)) ranging from $0.45$ to $0.83$, reflecting moderate to large differences.
Finally, while average \(\texttt{REC}\) across models is similar, their relationship with text length differs.
We observe that only models show increasing variation in referring expressions with longer text, e.g., \modelname{Gemini 2.0 Flash}.
Human stories remain consistent in how they refer to characters, regardless of length.

An interesting trend is that larger models often change characters more frequently than smaller ones, indicating weaker character consistency.
However, as Table~\ref{tab:generalstats} shows, they also generate longer outputs with more content, likely introducing more characters and switches.
In contrast, smaller models produce shorter, simpler outputs -- sometimes just a sentence or a word -- leading to lower CharCh scores that do not necessarily imply better coherence.
For instance, \modelname{DeepSeek-VL2-1B} often produces no mentions, while \modelname{DeepSeek-VL2-4.5B} generates many (Table~\ref{tab:generalstats2}).
Larger models tend to over-generate, resulting in dynamic but less grounded stories.

To support this, we report Pearson correlation between CharCh and MCC in Table~\ref{tab:corr}.
These consistently negative correlations (except in human stories) suggest that higher character turnover is linked to lower coherence.
Human-authored stories do not show this trend, suggesting that humans can manage frequent character changes without sacrificing clarity -- something models still struggle with.

\begin{table}[ht]
    \centering
    \begin{tabular}{l@{\hspace{0.25em}}r@{}l}
        \toprule
        \textbf{Model} & \multicolumn{2}{c}{\boldmath\(\rho\)} \\
        \midrule
        \modelname{DeepSeek-VL2-4.5B} & -0.236 & \textbf{**} \\
        \modelname{DeepSeek-VL2-1B}   & -0.263 & \textbf{**} \\
        \modelname{Gemini 2.0 Flash}  & -0.189 & \textbf{**} \\
        \modelname{GPT4o}             & -0.248 & \textbf{**} \\
        \modelname{InternVL2.5-78B}   & -0.251 & \textbf{**} \\
        \modelname{Qwen2-VL-72B}      & -0.238 & \textbf{**} \\
        \modelname{Qwen2-VL-7B}       & -0.188 & \textbf{**} \\
        \addlinespace[0.2em]
        \hline
        \addlinespace[0.2em]
        Human                         & -0.040 &           \\
        \bottomrule
    \end{tabular}
    \caption{Pearson correlation (\(\rho\)) between CharCh scores and MCC scores.
    Values marked with \textbf{**} are statistically significant (\(p < 0.001\)).}
    \label{tab:corr}
\end{table}

%\section{Discussion}

%\paragraph{Formal theories of discourse}
%Our work focuses on describing how character references in visual narratives behave, and how this can reflect coherence and tellability.
%While we do not directly build on formal discourse theories, we think that the concepts of information structure and the role of topic and focus together with insights from Centering Theory \citep{grosz-etal-1995-centering} are relevant.
%However, we chose not to rely on Centering Theory directly, as it imposes constraints such as assumptions about anaphora resolution \citep{lappin1994} and requires parameter tuning, including for concepts like ``utterance'' \citep{poesio2004} which are non-trivial to define.
%As shown by \citet{chai2022}, Centering Theory can complement but not replace modern neural coreference systems.
%Since our stories are grounded in image sequences, directly applying Centering Theory poses additional challenges that we hope to explore in future work.

%\paragraph{}
%While our current metrics focus on surface-level reference patterns, they still capture important aspects of tellability - such as continuity, salience, and character switching.
%Our contribution lies in showing that even basic coreference-based features can distinguish human and model narratives.
%Moreover, our MCC (Multimodal Character Continuity) metric bridges vision and language, going beyond pure text-level analysis. 

\vspace{-.1cm}
\section{Conclusions and future work}

We found substantial differences in coreferential patterns between LLM and human outputs.
The fact that humans subjectively perceive model outputs as human-like does not imply that the models actually behave in a human-like way.
Coreference helps humans structure narratives, and its divergence in LLMs has implications for AI-human interaction that require further exploration.
Our metrics allow us to measure the effects of the fact that, unlike humans, models are not explicitly required to caption each image.

The proposed metrics operate on textual references and can be applied to a range of formats, including dialogues, text-only stories \citep{fan-etal-2018-hierarchical}, and collaboratively written narratives \citep{akoury-etal-2020-storium}.
We plan to incorporate VIST \citep{huang-etal-2016-visual} and VIST-Character \citep{liu-detecting-2023}, which include detailed visual and textual coreference chains and importance ratings for characters, providing a strong basis for further evaluation of coreferential coherence.

Future work will explore LLM attention patterns to better understand biases in reference and coreference generation.
We are also actively considering interpretability experiments to probe how models internally represent characters during generation, for example, whether attention aligns with character mentions in images.
In addition, we aim to conduct human evaluations to better understand what makes a ``good'' story.

\section*{Limitations}

This work deals with the quality of generation in English.
In addition, presented metrics rely on the output of an automatic coreference system.
If a reliable model of coreference does not exist the metrics cannot be computed reliably.
While our main focus is on the analysis of coreferential patterns produced by recent multimodal models, we use data from only one visual storytelling dataset, VWP \citep{hong_visual_2023}.
Our metrics are also affected by the quality of automatically generated texts which we do not explicitly evaluate with automatic metrics or regenerate with different decoding methods.
We also note that prompt design can impact Instruction-following ability of the models which in turn can affect coherence of the generated stories.

\section*{Acknowledgments}
The work reported in this paper has been supported by a grant from the Swedish Research Council (VR project 2014-39) for the establishment of the Centre for Linguistic Theory and Studies in Probability (CLASP) at the University of Gothenburg.
The computations and data storage were enabled by resources provided by the National Academic Infrastructure for Supercomputing in Sweden (NAISS), partially funded by the Swedish Research Council through grant agreement no. 2022-06725.

\bibliography{gem2}

\appendix

\section{Multimodal character continuity: technical details}
\label{app:mmc_details}

We propose a metric for evaluating character continuity across visual and textual modalities in image-grounded stories.  
Our approach leverages existing character annotations from MovieNet \citep{huang2020movienetholisticdatasetmovie}, the dataset that also provides the images used in Visual Writing Prompts (VWP) \citep{hong_visual_2023}.  
In MovieNet, each image is associated with a movie and includes character detections labeled either with artificial names (e.g., \texttt{nm000004}) or as \texttt{Unknown}.
For each story \( X \) with an image sequence \( \mathcal{I} = \{i_1, i_2, \dots, i_K\} \), we process character annotations from MovieNet that include:  
(i) character bounding boxes \( B = \{b_1, b_2, \dots, b_n\} \) for each image,  
(ii) character identifiers (PIDs) mapped to actor names using MovieNet's cast metadata, and  
(iii) the relative size of bounding boxes (computed as a ratio to the image area).  
We then record the character names, effectively capturing which characters appear in which images.  
Mentions of characters in the texts are identified by matching these names to those in the annotations. e.g. ``russell'' in text is mapped with ``Russell'' in annotations.
Finally, \modelname{LinkAppend} allows us to determine when a character is mentioned in texts by analysing the coreference chains that include those character names.

\section{Model size and budget}
\label{sec:techdetails}

We used A100 40GB GPUs and A100 80GB GPUs to run models for our tasks (visual story generation and coreference resolution).
For story generation, the time required for the models took up to $14$ GPU hours.
For coreference resolution, the time required was up to $12$ GPU hours.
Closed models were prompted through their API.
We adjusted the prompts to the input format of each model according to their official documentation.
We used scripts from \url{https://github.com/boberle/corefconversion} to convert the output of \modelname{Link-Append} to appropriate format for our analysis (from .conll format to .jsonlines).

\section{Prompts}
\label{app:prompts}

Below are two texts that were given to the models for visual story generation.
The two texts differ slightly in phrasing depending on whether the data had images of characters from the story and their names.
The differences are highlighted in \textcolor{red}{red}.

\begin{tcolorbox}[
  float=t,
  colback=gray!10,
  colframe=gray!50,
  sharp corners,
  boxrule=1pt,
  title={Prompt text A},
  width=\columnwidth
]
\textcolor{red}{View a sequence of N images and figure out the content. Then write a story with it.} View a sequence of images as many times as you wish. Figure out who were involved and what happened. Then write a story that fits the image sequence. You should write the story using at least 5 images. You need to write at least 50 but no more than 300 words. You do not need to write text without a corresponding image unless it is necessary. The story should be related to the image sequence. Describe only the most important character(s) and event(s). You can use either the first name, a pronoun, or a noun phrase according to the context. If the character you want to mention is not there, name the characters as you want, but please be consistent. Use punctuation and letter case correctly. Do not mention that you are describing an image. Avoid using phrases like ``In this image, ...''. Do not write a monologue of a character or a dialogue between characters.
\end{tcolorbox}

\begin{tcolorbox}[
  float=t,
  colback=gray!10,
  colframe=gray!50,
  sharp corners,
  boxrule=1pt,
  title={Prompt text B},
  width=\columnwidth
]
\textcolor{red}{View a sequence of N images followed by K character images and figure out the content.} Then write a story with it. View a sequence of images as many times as you wish. Figure out who were involved and what happened. Then write a story that fits the image sequence. You should write the story using at least 5 images. You need to write at least 50 but no more than 300 words. You do not need to write text without a corresponding image unless it is necessary. The story should be related to the image sequence. Describe only the most important character(s) and event(s). \textcolor{red}{When mentioning the characters, please follow their names which are provided in the order that the character images were given: [character names].} You can use either the first name, a pronoun, or a noun phrase according to the context. If the character you want to mention is not there, name the characters as you want, but please be consistent. Use punctuation and letter case correctly. Do not mention that you are describing an image. Avoid using phrases like ``In this image, ...''. Do not write a monologue of a character or a dialogue between characters.
\end{tcolorbox}

\section{Dataset distribution}
\label{sec:data_distr}

\begin{figure}[ht]
    \centering
    \begin{subfigure}[b]{0.5\linewidth}
        \centering
        \includegraphics[width=\linewidth]{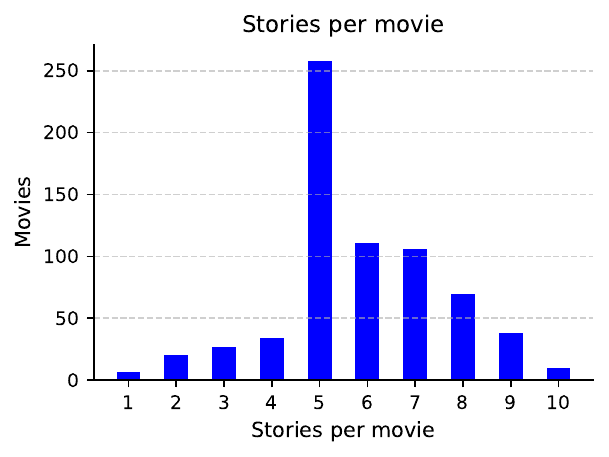}\label{subfig:spm}
        \end{subfigure}  
    \hfill
    \begin{subfigure}[b]{0.45\linewidth}
        \centering
        \includegraphics[width=\linewidth]{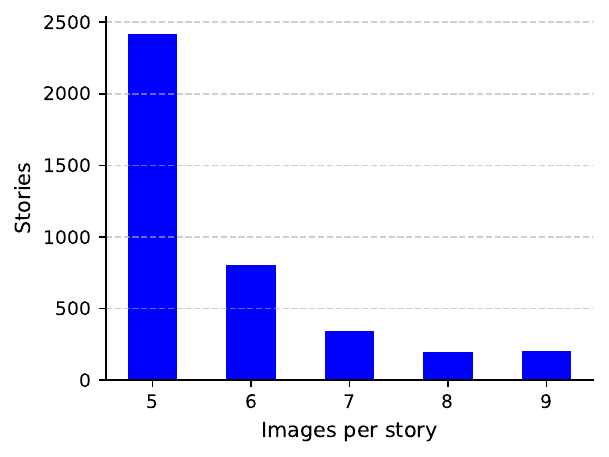}\label{subfig:ipm}
    \end{subfigure}
        \caption{Distributions in our evaluation dataset.}\label{fig:eval-set}
    \label{fig:distrdistr}
\end{figure}

A general distribution of visual stories in our dataset in terms of movies and number of images is shown in Figure~\ref{fig:distrdistr}.

\onecolumn

\noindent
\begin{table*}[ht!]
\section{Example I}
\label{sec:examples}

Below you see a sequence of images depicting a visual story alongside bounding boxes of characters that were given to models and humans.
Next, we show texts produced by humans and models, including coreference chains produced by \modelname{Link-Append}.
In this example human text has one of the highest CharTr values across all examples in the dataset ($2.33$).
We provide corresponding results for CharTr metric for each model alongside text that they generated.

\vspace{.5cm}

\begin{tabular}{cc}
  \shortstack{\textbf{Image 1}\\ \includegraphics[width=0.48\textwidth]{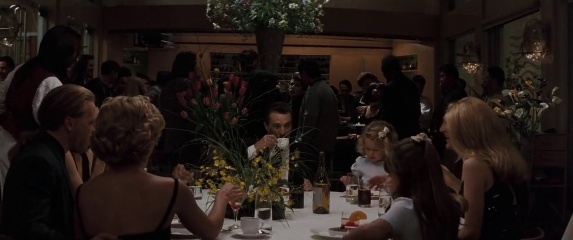}} &
  \shortstack{\textbf{Image 2}\\ \includegraphics[width=0.48\textwidth]{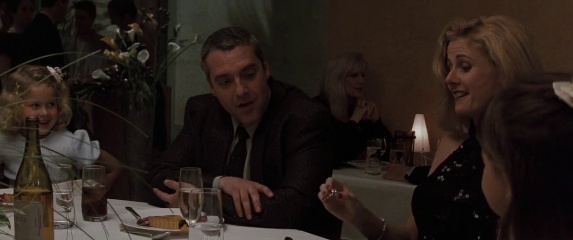}} \\[6pt]
  \shortstack{\textbf{Image 3}\\ \includegraphics[width=0.48\textwidth]{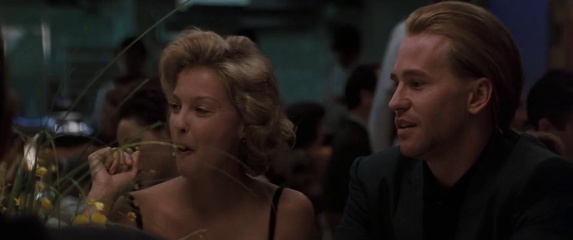}} &
  \shortstack{\textbf{Image 4}\\ \includegraphics[width=0.48\textwidth]{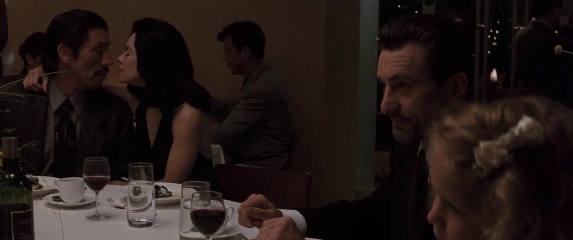}} \\[6pt]
  \multicolumn{2}{c}{%
    \shortstack{\textbf{Image 5}\\ \includegraphics[width=0.48\textwidth]{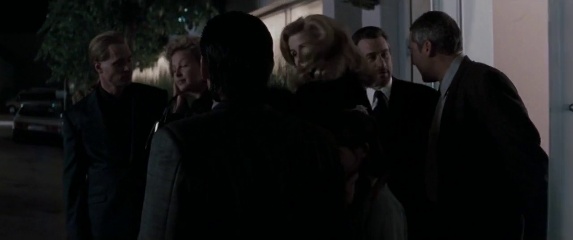}}
  }\\
\end{tabular}

\begin{center}
\begin{tabular}{ccc}
\centering
  \shortstack{\textbf{Character image 1: Ashley}\\ \includegraphics[width=0.20\textwidth]{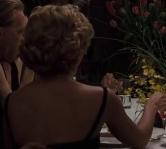}} &
  \shortstack{\textbf{Character image 2: Val}\\ \includegraphics[width=0.14\textwidth]{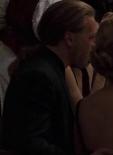}} &
  \shortstack{\textbf{Character image 3: Tom}\\ \includegraphics[width=0.20\textwidth]{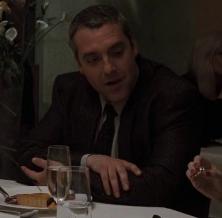}} \\
  \shortstack{\textbf{Character image 4: Robert}\\ \includegraphics[width=0.20\textwidth]{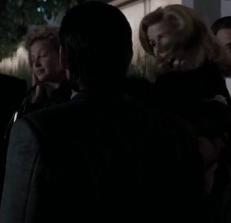}} &
    \shortstack{\textbf{Character image 5: Amy}\\ \includegraphics[width=0.20\textwidth]{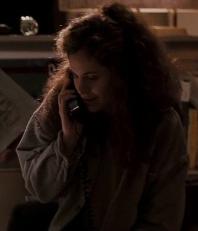}
  }\\
\end{tabular}
\end{center}
\end{table*}

\vspace{-.5cm}

\begin{tcolorbox}[enhanced, breakable]
            \footnotesize
            \vspace{.5cm}
            \textbf{Human, CharTr $2.33$:}\\

\textcolor{orange}{[}\textcolor{red}{[}Ashley\textcolor{red}{]}$^{\textcolor{red}{1}}$ and \textcolor{blue}{[}Val\textcolor{blue}{]}$^{\textcolor{blue}{2}}$\textcolor{orange}{]}$^{\textcolor{orange}{3}}$ are sitting together at \textcolor{teal}{[}a table\textcolor{teal}{]}$^{\textcolor{teal}{4}}$ at \textcolor{purple}{[}the Gala\textcolor{purple}{]}$^{\textcolor{purple}{5}}$ .

\textcolor{brown}{[}Tom\textcolor{brown}{]}$^{\textcolor{brown}{6}}$ was sitting across from \textcolor{orange}{[}them\textcolor{orange}{]}$^{\textcolor{orange}{3}}$ , and \textcolor{brown}{[}Tom\textcolor{brown}{]}$^{\textcolor{brown}{6}}$ leans in to ask what \textcolor{orange}{[}their\textcolor{orange}{]}$^{\textcolor{orange}{3}}$ secret to a loving and long marriage is .

Flattered , \textcolor{orange}{[}\textcolor{red}{[}Ashley\textcolor{red}{]}$^{\textcolor{red}{1}}$ and \textcolor{blue}{[}Val\textcolor{blue}{]}$^{\textcolor{blue}{2}}$\textcolor{orange}{]}$^{\textcolor{orange}{3}}$ smile .

\textcolor{red}{[}Ashley\textcolor{red}{]}$^{\textcolor{red}{1}}$ starts to tell \textcolor{brown}{[}Tom\textcolor{brown}{]}$^{\textcolor{brown}{6}}$ that \textcolor{blue}{[}Val\textcolor{blue}{]}$^{\textcolor{blue}{2}}$ does everything \textcolor{red}{[}she\textcolor{red}{]}$^{\textcolor{red}{1}}$ tells \textcolor{blue}{[}him\textcolor{blue}{]}$^{\textcolor{blue}{2}}$ to .

\textcolor{magenta}{[}Robert\textcolor{magenta}{]}$^{\textcolor{magenta}{7}}$ got stuck sitting at the kids ' table at \textcolor{purple}{[}the Gala\textcolor{purple}{]}$^{\textcolor{purple}{5}}$ , but \textcolor{magenta}{[}he\textcolor{magenta}{]}$^{\textcolor{magenta}{7}}$ is starting at \textcolor{teal}{[}\textcolor{orange}{[}\textcolor{blue}{[}Val\textcolor{blue}{]}$^{\textcolor{blue}{2}}$ and \textcolor{red}{[}Ashley\textcolor{red}{]}$^{\textcolor{red}{1}}$ 's\textcolor{orange}{]}$^{\textcolor{orange}{3}}$ table\textcolor{teal}{]}$^{\textcolor{teal}{4}}$ while wishing \textcolor{magenta}{[}he\textcolor{magenta}{]}$^{\textcolor{magenta}{7}}$ could be sitting there with \textcolor{orange}{[}them\textcolor{orange}{]}$^{\textcolor{orange}{3}}$ .

After \textcolor{purple}{[}the Gala\textcolor{purple}{]}$^{\textcolor{purple}{5}}$ , \textcolor{olive}{[}\textcolor{blue}{[}Val\textcolor{blue}{]}$^{\textcolor{blue}{2}}$ , \textcolor{red}{[}Ashley\textcolor{red}{]}$^{\textcolor{red}{1}}$ , and everyone at \textcolor{teal}{[}\textcolor{olive}{[}their\textcolor{olive}{]}$^{\textcolor{olive}{8}}$ table\textcolor{teal}{]}$^{\textcolor{teal}{4}}$\textcolor{olive}{]}$^{\textcolor{olive}{8}}$ are standing outside the hotel discussing when and where \textcolor{olive}{[}their\textcolor{olive}{]}$^{\textcolor{olive}{8}}$ next party will be .

\textcolor{magenta}{[}Robert\textcolor{magenta}{]}$^{\textcolor{magenta}{7}}$ is walking up behind \textcolor{olive}{[}them\textcolor{olive}{]}$^{\textcolor{olive}{8}}$ to join \textcolor{olive}{[}their\textcolor{olive}{]}$^{\textcolor{olive}{8}}$ conversation .

        \vspace{.5cm}
        \textbf{\modelname{DeepSeekVL2-1B}, CharTr $0.46$:}\\

\textcolor{red}{[}The sequence of images\textcolor{red}{]}$^{\textcolor{red}{1}}$ depicts \textcolor{blue}{[}a group of people gathered at a dinner table in what appears to be a restaurant setting\textcolor{blue}{]}$^{\textcolor{blue}{2}}$ .

The atmosphere is warm and intimate , with soft lighting and floral centerpieces adding to the ambiance .

\textcolor{blue}{[}The individuals\textcolor{blue}{]}$^{\textcolor{blue}{2}}$ are engaged in conversation , some smiling and others looking at each other , suggesting a lively and friendly gathering .

\textcolor{teal}{[}Ashley , Val , Tom , Robert , and \textcolor{orange}{[}Amy\textcolor{orange}{]}$^{\textcolor{orange}{3}}$\textcolor{teal}{]}$^{\textcolor{teal}{4}}$ are the main characters in \textcolor{purple}{[}this scene\textcolor{purple}{]}$^{\textcolor{purple}{5}}$ .

\textcolor{teal}{[}They\textcolor{teal}{]}$^{\textcolor{teal}{4}}$ are seated around a table , with plates of food and drinks in front of \textcolor{teal}{[}them\textcolor{teal}{]}$^{\textcolor{teal}{4}}$ .

The setting is indoors , likely in a restaurant , as indicated by the decor and lighting .

The background shows other diners and tables , creating a bustling environment .

The story that fits \textcolor{red}{[}the sequence of images\textcolor{red}{]}$^{\textcolor{red}{1}}$ is as follows : \textcolor{teal}{[}\textcolor{brown}{[}Ashley\textcolor{brown}{]}$^{\textcolor{brown}{6}}$ , \textcolor{magenta}{[}Val\textcolor{magenta}{]}$^{\textcolor{magenta}{7}}$ , \textcolor{olive}{[}Tom\textcolor{olive}{]}$^{\textcolor{olive}{8}}$ , \textcolor{cyan}{[}Robert\textcolor{cyan}{]}$^{\textcolor{cyan}{9}}$ , and \textcolor{orange}{[}Amy\textcolor{orange}{]}$^{\textcolor{orange}{3}}$\textcolor{teal}{]}$^{\textcolor{teal}{4}}$ are friends who have gathered for a dinner party .

\textcolor{teal}{[}They\textcolor{teal}{]}$^{\textcolor{teal}{4}}$ are enjoying each other 's company , sharing stories and laughter .

The atmosphere is relaxed and convivial , with everyone feeling comfortable and at ease .

The presence of wine and the elegant table setting suggest that this is a special occasion , perhaps a birthday or anniversary dinner .

\textcolor{teal}{[}The characters\textcolor{teal}{]}$^{\textcolor{teal}{4}}$ are engaged in conversation , with some looking at each other and \textcolor{lime}{[}others\textcolor{lime}{]}$^{\textcolor{lime}{10}}$ focused on \textcolor{lime}{[}their\textcolor{lime}{]}$^{\textcolor{lime}{10}}$ food .

\textcolor{purple}{[}The scene\textcolor{purple}{]}$^{\textcolor{purple}{5}}$ captures a moment of connection and camaraderie among \textcolor{teal}{[}the group\textcolor{teal}{]}$^{\textcolor{teal}{4}}$ .

        \vspace{.5cm}
        \textbf{\modelname{DeepSeek-VL2-4.5B}, CharTr $0.66$:}\\

The evening began with a sense of anticipation as \textcolor{red}{[}the group of six\textcolor{red}{]}$^{\textcolor{red}{1}}$ gathered around \textcolor{blue}{[}the table in \textcolor{orange}{[}the dimly lit restaurant\textcolor{orange}{]}$^{\textcolor{orange}{3}}$\textcolor{blue}{]}$^{\textcolor{blue}{2}}$ .

\textcolor{teal}{[}Ashley , the elegant woman with blonde hair\textcolor{teal}{]}$^{\textcolor{teal}{4}}$ , sat at the head of \textcolor{blue}{[}the table\textcolor{blue}{]}$^{\textcolor{blue}{2}}$ , \textcolor{teal}{[}her\textcolor{teal}{]}$^{\textcolor{teal}{4}}$ eyes scanning \textcolor{orange}{[}the room\textcolor{orange}{]}$^{\textcolor{orange}{3}}$ with a mixture of curiosity and caution .

\textcolor{purple}{[}Val , the man in the dark suit\textcolor{purple}{]}$^{\textcolor{purple}{5}}$ , sat beside \textcolor{teal}{[}her\textcolor{teal}{]}$^{\textcolor{teal}{4}}$ , \textcolor{purple}{[}his\textcolor{purple}{]}$^{\textcolor{purple}{5}}$ hands clasped together as \textcolor{purple}{[}he\textcolor{purple}{]}$^{\textcolor{purple}{5}}$ listened intently to \textcolor{brown}{[}the conversation\textcolor{brown}{]}$^{\textcolor{brown}{6}}$ .

Across from them , \textcolor{magenta}{[}Tom , the man with the neatly combed hair\textcolor{magenta}{]}$^{\textcolor{magenta}{7}}$ , leaned forward , \textcolor{magenta}{[}his\textcolor{magenta}{]}$^{\textcolor{magenta}{7}}$ eyes locked onto \textcolor{olive}{[}Robert , the man with the serious expression , who was deep in conversation with \textcolor{cyan}{[}Amy , the woman with the dark hair\textcolor{cyan}{]}$^{\textcolor{cyan}{9}}$\textcolor{olive}{]}$^{\textcolor{olive}{8}}$ .

As \textcolor{lime}{[}the night\textcolor{lime}{]}$^{\textcolor{lime}{10}}$ wore on , \textcolor{brown}{[}\textcolor{red}{[}the group 's\textcolor{red}{]}$^{\textcolor{red}{1}}$ conversation\textcolor{brown}{]}$^{\textcolor{brown}{6}}$ turned more serious .

\textcolor{olive}{[}Robert 's\textcolor{olive}{]}$^{\textcolor{olive}{8}}$ words were laced with urgency , and \textcolor{cyan}{[}Amy 's\textcolor{cyan}{]}$^{\textcolor{cyan}{9}}$ eyes reflected a mix of concern and determination .

\textcolor{magenta}{[}Tom , ever the observer\textcolor{magenta}{]}$^{\textcolor{magenta}{7}}$ , noticed the tension in the air and tried to steer \textcolor{brown}{[}the conversation\textcolor{brown}{]}$^{\textcolor{brown}{6}}$ towards lighter topics .

\textcolor{purple}{[}Val\textcolor{purple}{]}$^{\textcolor{purple}{5}}$ , however , remained focused on the matter at hand , \textcolor{purple}{[}his\textcolor{purple}{]}$^{\textcolor{purple}{5}}$ mind racing with possibilities .

Outside , \textcolor{lime}{[}the night\textcolor{lime}{]}$^{\textcolor{lime}{10}}$ was quiet , the only sound the distant hum of the city .

\textcolor{brown}{[}\textcolor{red}{[}The group 's\textcolor{red}{]}$^{\textcolor{red}{1}}$ discussion\textcolor{brown}{]}$^{\textcolor{brown}{6}}$ had reached a critical point , and the stakes were high .

As \textcolor{red}{[}they\textcolor{red}{]}$^{\textcolor{red}{1}}$ prepared to leave , \textcolor{red}{[}the group 's\textcolor{red}{]}$^{\textcolor{red}{1}}$ expressions were a mix of relief and resolve .

\textcolor{red}{[}They\textcolor{red}{]}$^{\textcolor{red}{1}}$ knew that the decisions \textcolor{red}{[}they\textcolor{red}{]}$^{\textcolor{red}{1}}$ made that night would have far-reaching consequences .

As \textcolor{red}{[}they\textcolor{red}{]}$^{\textcolor{red}{1}}$ stepped into \textcolor{lime}{[}the night\textcolor{lime}{]}$^{\textcolor{lime}{10}}$ , \textcolor{red}{[}the group 's\textcolor{red}{]}$^{\textcolor{red}{1}}$ unity was palpable .

\textcolor{red}{[}They\textcolor{red}{]}$^{\textcolor{red}{1}}$ moved as one , \textcolor{red}{[}their\textcolor{red}{]}$^{\textcolor{red}{1}}$ purpose clear .

The city lights flickered around \textcolor{red}{[}them\textcolor{red}{]}$^{\textcolor{red}{1}}$ , casting long shadows that seemed to stretch into the future .

\textcolor{lime}{[}The night\textcolor{lime}{]}$^{\textcolor{lime}{10}}$ had brought \textcolor{red}{[}them\textcolor{red}{]}$^{\textcolor{red}{1}}$ together , and now , \textcolor{red}{[}they\textcolor{red}{]}$^{\textcolor{red}{1}}$ were ready to face whatever challenges lay ahead .

        \vspace{.5cm}
        \textbf{\modelname{Gemini 2.0 Flash}, CharTr None:}\\

        Val and Tom attend a dinner party at a local restaurant .

        \vspace{.5cm}
        \textbf{\modelname{GPT4o}, CharTr $0.75$:}\\

\textcolor{brown}{[}\textcolor{red}{[}Ashley\textcolor{red}{]}$^{\textcolor{red}{1}}$ , \textcolor{blue}{[}Val\textcolor{blue}{]}$^{\textcolor{blue}{2}}$ , \textcolor{orange}{[}Tom\textcolor{orange}{]}$^{\textcolor{orange}{3}}$ , \textcolor{teal}{[}Robert\textcolor{teal}{]}$^{\textcolor{teal}{4}}$ , and \textcolor{purple}{[}Amy\textcolor{purple}{]}$^{\textcolor{purple}{5}}$\textcolor{brown}{]}$^{\textcolor{brown}{6}}$ gathered at an elegant restaurant to celebrate a special occasion .

The atmosphere was lively , with \textcolor{brown}{[}everyone\textcolor{brown}{]}$^{\textcolor{brown}{6}}$ enjoying the company and the food .

\textcolor{red}{[}Ashley\textcolor{red}{]}$^{\textcolor{red}{1}}$ and \textcolor{blue}{[}Val\textcolor{blue}{]}$^{\textcolor{blue}{2}}$ , sitting closely , shared a laugh over a shared memory , creating a warm and joyful mood at \textcolor{magenta}{[}the table\textcolor{magenta}{]}$^{\textcolor{magenta}{7}}$ .

\textcolor{orange}{[}Tom\textcolor{orange}{]}$^{\textcolor{orange}{3}}$ , engaged in conversation with \textcolor{teal}{[}Robert\textcolor{teal}{]}$^{\textcolor{teal}{4}}$ , gestured animatedly , drawing laughter from both .

Across \textcolor{magenta}{[}the table\textcolor{magenta}{]}$^{\textcolor{magenta}{7}}$ , \textcolor{purple}{[}Amy\textcolor{purple}{]}$^{\textcolor{purple}{5}}$ was sharing a story that held \textcolor{brown}{[}everyone 's\textcolor{brown}{]}$^{\textcolor{brown}{6}}$ attention , \textcolor{purple}{[}her\textcolor{purple}{]}$^{\textcolor{purple}{5}}$ expressions enchanting \textcolor{brown}{[}the group\textcolor{brown}{]}$^{\textcolor{brown}{6}}$ .

After dinner , \textcolor{brown}{[}they \textcolor{brown}{[}all\textcolor{brown}{]}$^{\textcolor{brown}{6}}$\textcolor{brown}{]}$^{\textcolor{brown}{6}}$ decided to step outside for a breath of fresh air .

The cool night air filled with \textcolor{brown}{[}their\textcolor{brown}{]}$^{\textcolor{brown}{6}}$ lively chatter created a serene end to \textcolor{olive}{[}a wonderful evening\textcolor{olive}{]}$^{\textcolor{olive}{8}}$ .

As \textcolor{brown}{[}they\textcolor{brown}{]}$^{\textcolor{brown}{6}}$ stood together , \textcolor{brown}{[}they\textcolor{brown}{]}$^{\textcolor{brown}{6}}$ realized the depth of \textcolor{brown}{[}their\textcolor{brown}{]}$^{\textcolor{brown}{6}}$ camaraderie .

Meanwhile , \textcolor{purple}{[}Amy\textcolor{purple}{]}$^{\textcolor{purple}{5}}$ received an unexpected phone call that seemed urgent .

\textcolor{purple}{[}Her\textcolor{purple}{]}$^{\textcolor{purple}{5}}$ expression turned serious , reminding \textcolor{purple}{[}her\textcolor{purple}{]}$^{\textcolor{purple}{5}}$ of the responsibilities awaiting \textcolor{purple}{[}her\textcolor{purple}{]}$^{\textcolor{purple}{5}}$ return .

\textcolor{purple}{[}She\textcolor{purple}{]}$^{\textcolor{purple}{5}}$ promised to meet \textcolor{brown}{[}the others\textcolor{brown}{]}$^{\textcolor{brown}{6}}$ again soon , cherishing \textcolor{olive}{[}the evening 's\textcolor{olive}{]}$^{\textcolor{olive}{8}}$ joyful moments .

\textcolor{brown}{[}Their\textcolor{brown}{]}$^{\textcolor{brown}{6}}$ gathering ended with warm goodbyes , \textcolor{brown}{[}everyone\textcolor{brown}{]}$^{\textcolor{brown}{6}}$ cherishing the memories \textcolor{brown}{[}they\textcolor{brown}{]}$^{\textcolor{brown}{6}}$ had made together .

The shared laughter and stories would linger , a testament to \textcolor{brown}{[}their\textcolor{brown}{]}$^{\textcolor{brown}{6}}$ enduring friendship .

        \vspace{.5cm}
        \textbf{\modelname{InternVL2.5-78B}, CharTr $0.64$:}\\

It was a chilly evening in the small town of Millfield , where \textcolor{red}{[}a formal banquet\textcolor{red}{]}$^{\textcolor{red}{1}}$ had gathered a group of old friends for \textcolor{blue}{[}a reunion at \textcolor{orange}{[}a luxurious restaurant\textcolor{orange}{]}$^{\textcolor{orange}{3}}$\textcolor{blue}{]}$^{\textcolor{blue}{2}}$ .

As \textcolor{teal}{[}guests\textcolor{teal}{]}$^{\textcolor{teal}{4}}$ mingled near \textcolor{teal}{[}their\textcolor{teal}{]}$^{\textcolor{teal}{4}}$ tables , \textcolor{brown}{[}\textcolor{purple}{[}Ashley\textcolor{purple}{]}$^{\textcolor{purple}{5}}$ and Robert\textcolor{brown}{]}$^{\textcolor{brown}{6}}$ were seated at the center of \textcolor{red}{[}the banquet\textcolor{red}{]}$^{\textcolor{red}{1}}$ , engaged in a lively conversation .

The table was adorned with a vase of vibrant blossoms , and the air was filled with the hum of laughter and clinking glasses .

In the background , \textcolor{cyan}{[}\textcolor{magenta}{[}Val\textcolor{magenta}{]}$^{\textcolor{magenta}{7}}$ and \textcolor{olive}{[}Tom\textcolor{olive}{]}$^{\textcolor{olive}{8}}$\textcolor{cyan}{]}$^{\textcolor{cyan}{9}}$ were deeply engrossed in \textcolor{cyan}{[}their\textcolor{cyan}{]}$^{\textcolor{cyan}{9}}$ conversation , sharing stories and memories of \textcolor{cyan}{[}their\textcolor{cyan}{]}$^{\textcolor{cyan}{9}}$ youth .

\textcolor{cyan}{[}Their\textcolor{cyan}{]}$^{\textcolor{cyan}{9}}$ affectionate exchanges brought warmth to \textcolor{cyan}{[}their\textcolor{cyan}{]}$^{\textcolor{cyan}{9}}$ surroundings , eliciting gentle smiles from fellow guests .

As \textcolor{lime}{[}the evening\textcolor{lime}{]}$^{\textcolor{lime}{10}}$ unfolded , the atmosphere was filled with camaraderie and joyous reunion .

\textcolor{brown}{[}\textcolor{purple}{[}Ashley\textcolor{purple}{]}$^{\textcolor{purple}{5}}$ and \textcolor{magenta}{[}Val\textcolor{magenta}{]}$^{\textcolor{magenta}{7}}$\textcolor{brown}{]}$^{\textcolor{brown}{6}}$ shared light-hearted banter with the other guests around \textcolor{brown}{[}their\textcolor{brown}{]}$^{\textcolor{brown}{6}}$ table .

The conversation was animated , as \textcolor{brown}{[}they\textcolor{brown}{]}$^{\textcolor{brown}{6}}$ reminisced on shared experiences and adventures from the past .

In a quiet corner , an unexpected moment caught everyone ’ s attention .

\textcolor{cyan}{[}\textcolor{magenta}{[}Val\textcolor{magenta}{]}$^{\textcolor{magenta}{7}}$ and \textcolor{olive}{[}Tom\textcolor{olive}{]}$^{\textcolor{olive}{8}}$\textcolor{cyan}{]}$^{\textcolor{cyan}{9}}$ were tenderly holding hands , \textcolor{cyan}{[}their\textcolor{cyan}{]}$^{\textcolor{cyan}{9}}$ eyes meeting with a love so pure and unmarred by the passage of time .

This tender display of affection brought a wave of warmth even to those who had merely observed from afar .

As night deepened , \textcolor{blue}{[}the gathering\textcolor{blue}{]}$^{\textcolor{blue}{2}}$ began to wind down , but the heartfelt connections and cherished moments lingered in the air long after \textcolor{teal}{[}the guests\textcolor{teal}{]}$^{\textcolor{teal}{4}}$ had bid \textcolor{teal}{[}their\textcolor{teal}{]}$^{\textcolor{teal}{4}}$ farewells and left \textcolor{orange}{[}the restaurant 's\textcolor{orange}{]}$^{\textcolor{orange}{3}}$ warm embrace .

\textcolor{lime}{[}The evening\textcolor{lime}{]}$^{\textcolor{lime}{10}}$ had been a poignant reminder of the power of friendship and the enduring bonds that time can not extinguish .

        \vspace{.5cm}
        \textbf{\modelname{Qwen2-VL-72B}, CharTr $0.17$:}\\

\textcolor{red}{[}The evening\textcolor{red}{]}$^{\textcolor{red}{1}}$ began with a lively gathering at \textcolor{orange}{[}a restaurant , where \textcolor{blue}{[}friends and family\textcolor{blue}{]}$^{\textcolor{blue}{2}}$ sat around a table adorned with flowers and enjoyed each other 's company\textcolor{orange}{]}$^{\textcolor{orange}{3}}$ .

\textcolor{teal}{[}Ashley\textcolor{teal}{]}$^{\textcolor{teal}{4}}$ , dressed elegantly , engaged in animated conversation with \textcolor{teal}{[}her\textcolor{teal}{]}$^{\textcolor{teal}{4}}$ companions , while Val and Tom listened intently .

As the night progressed , \textcolor{blue}{[}the group\textcolor{blue}{]}$^{\textcolor{blue}{2}}$ decided to move on to another location for further celebrations .

Outside \textcolor{orange}{[}the restaurant\textcolor{orange}{]}$^{\textcolor{orange}{3}}$ , \textcolor{blue}{[}they\textcolor{blue}{]}$^{\textcolor{blue}{2}}$ formed a small circle , continuing \textcolor{blue}{[}their\textcolor{blue}{]}$^{\textcolor{blue}{2}}$ discussions .

Amy , who had been busy earlier , joined \textcolor{blue}{[}them\textcolor{blue}{]}$^{\textcolor{blue}{2}}$ later , bringing along a cheerful energy .

\textcolor{blue}{[}The group\textcolor{blue}{]}$^{\textcolor{blue}{2}}$ laughed and shared stories , creating a warm atmosphere despite the cool evening air .

Meanwhile , \textcolor{purple}{[}Robert , who was not part of the initial gathering\textcolor{purple}{]}$^{\textcolor{purple}{5}}$ , found \textcolor{purple}{[}himself\textcolor{purple}{]}$^{\textcolor{purple}{5}}$ in a different setting , engrossed in a phone call .

\textcolor{purple}{[}His\textcolor{purple}{]}$^{\textcolor{purple}{5}}$ expression suggested \textcolor{purple}{[}he\textcolor{purple}{]}$^{\textcolor{purple}{5}}$ was deep in thought , possibly dealing with some pressing matter .

Unbeknownst to \textcolor{purple}{[}him\textcolor{purple}{]}$^{\textcolor{purple}{5}}$ , \textcolor{purple}{[}his\textcolor{purple}{]}$^{\textcolor{purple}{5}}$ path would soon cross with \textcolor{blue}{[}the lively group\textcolor{blue}{]}$^{\textcolor{blue}{2}}$ , adding a new dynamic to \textcolor{red}{[}the evening 's\textcolor{red}{]}$^{\textcolor{red}{1}}$ events .

        \vspace{.5cm}
        \textbf{\modelname{Qwen2-VL-7B}, CharTr $1.29$:}\\

\textcolor{red}{[}The story\textcolor{red}{]}$^{\textcolor{red}{1}}$ revolves around \textcolor{blue}{[}a group of people who gather for a dinner party at a restaurant\textcolor{blue}{]}$^{\textcolor{blue}{2}}$ .

Among \textcolor{blue}{[}them\textcolor{blue}{]}$^{\textcolor{blue}{2}}$ are \textcolor{orange}{[}Ashley\textcolor{orange}{]}$^{\textcolor{orange}{3}}$ , \textcolor{teal}{[}Val\textcolor{teal}{]}$^{\textcolor{teal}{4}}$ , \textcolor{purple}{[}Tom\textcolor{purple}{]}$^{\textcolor{purple}{5}}$ , \textcolor{brown}{[}Robert\textcolor{brown}{]}$^{\textcolor{brown}{6}}$ , and \textcolor{magenta}{[}Amy\textcolor{magenta}{]}$^{\textcolor{magenta}{7}}$ .

The scene begins with \textcolor{blue}{[}a large group of people seated at \textcolor{olive}{[}a long table\textcolor{olive}{]}$^{\textcolor{olive}{8}}$ , engaged in lively conversation and enjoying \textcolor{blue}{[}their\textcolor{blue}{]}$^{\textcolor{blue}{2}}$ meal\textcolor{blue}{]}$^{\textcolor{blue}{2}}$ .

As the evening progresses , the atmosphere becomes more intimate , with smaller groups forming around \textcolor{olive}{[}the table\textcolor{olive}{]}$^{\textcolor{olive}{8}}$ .

In one of the images , \textcolor{cyan}{[}\textcolor{purple}{[}Tom\textcolor{purple}{]}$^{\textcolor{purple}{5}}$ and \textcolor{brown}{[}Robert\textcolor{brown}{]}$^{\textcolor{brown}{6}}$\textcolor{cyan}{]}$^{\textcolor{cyan}{9}}$ are seen in a quiet conversation , suggesting a deeper connection between \textcolor{cyan}{[}them\textcolor{cyan}{]}$^{\textcolor{cyan}{9}}$ .

Meanwhile , \textcolor{orange}{[}Ashley\textcolor{orange}{]}$^{\textcolor{orange}{3}}$ and \textcolor{teal}{[}Val\textcolor{teal}{]}$^{\textcolor{teal}{4}}$ are seated together , sharing a moment of laughter and camaraderie .

\textcolor{magenta}{[}Amy\textcolor{magenta}{]}$^{\textcolor{magenta}{7}}$ , on the other hand , is seen in a more solitary moment , talking on the phone , perhaps discussing something important or personal .

As the night draws to a close , \textcolor{blue}{[}the group\textcolor{blue}{]}$^{\textcolor{blue}{2}}$ begins to disperse .

\textcolor{cyan}{[}\textcolor{brown}{[}Robert\textcolor{brown}{]}$^{\textcolor{brown}{6}}$ and \textcolor{purple}{[}Tom\textcolor{purple}{]}$^{\textcolor{purple}{5}}$\textcolor{cyan}{]}$^{\textcolor{cyan}{9}}$ are seen in a final conversation , perhaps reflecting on \textcolor{lime}{[}the evening\textcolor{lime}{]}$^{\textcolor{lime}{10}}$ or discussing something of mutual interest .

\textcolor{magenta}{[}Amy\textcolor{magenta}{]}$^{\textcolor{magenta}{7}}$ is also seen in a moment of solitude , deep in thought , possibly contemplating the events of \textcolor{lime}{[}the evening\textcolor{lime}{]}$^{\textcolor{lime}{10}}$ .

\textcolor{red}{[}The story\textcolor{red}{]}$^{\textcolor{red}{1}}$ is one of camaraderie , reflection , and the passage of time , as \textcolor{blue}{[}the group of friends and acquaintances\textcolor{blue}{]}$^{\textcolor{blue}{2}}$ gather , share stories , and then part ways , each with \textcolor{blue}{[}their\textcolor{blue}{]}$^{\textcolor{blue}{2}}$ own thoughts and memories of \textcolor{lime}{[}the evening\textcolor{lime}{]}$^{\textcolor{lime}{10}}$ .

\end{tcolorbox}
%\clearpage

\noindent
\begin{table*}[ht]
\section{Example II}
\label{sec:examples2}

In this story example we have one of the highest multimodal character continuity (MCC) scores in the dataset for human-generated texts.
The scores for both humans and models are provided alongside generated texts.

\vspace{.5cm}

\begin{tabular}{cc}
  \shortstack{\textbf{Image 1}\\ \includegraphics[width=0.48\textwidth]{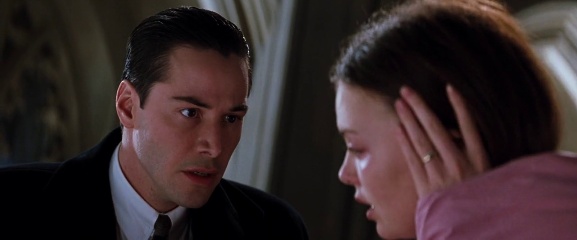}} &
  \shortstack{\textbf{Image 2}\\ \includegraphics[width=0.48\textwidth]{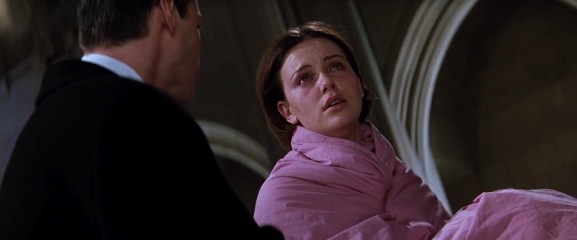}} \\[6pt]
  \shortstack{\textbf{Image 3}\\ \includegraphics[width=0.48\textwidth]{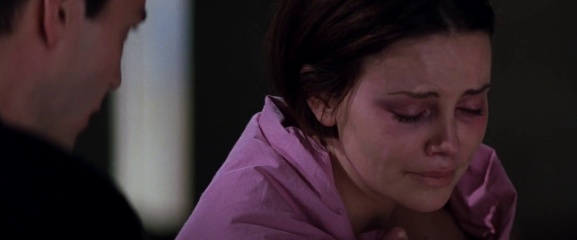}} &
  \shortstack{\textbf{Image 4}\\ \includegraphics[width=0.48\textwidth]{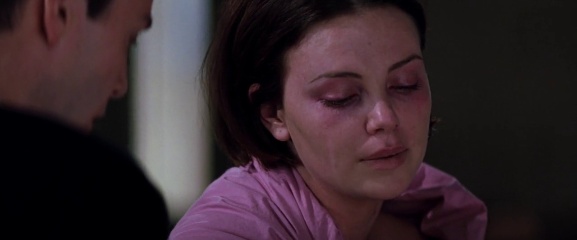}} \\[6pt]
  \multicolumn{2}{c}{%
    \shortstack{\textbf{Image 5}\\ \includegraphics[width=0.48\textwidth]{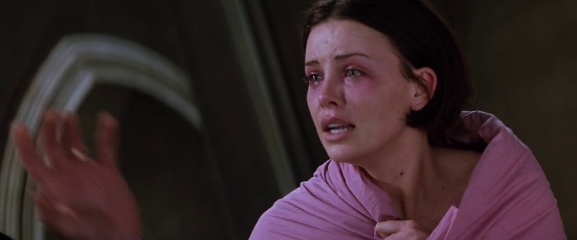}}
  }\\
\end{tabular}

\begin{center}
\begin{tabular}{cc}
\centering
  \shortstack{\textbf{Character image 1: Keanu}\\ \includegraphics[width=0.20\textwidth]{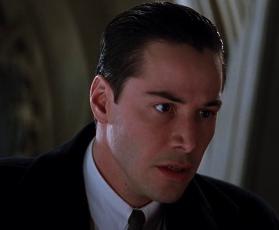}} &
  \shortstack{\textbf{Character image 2: Charlize}\\ \includegraphics[width=0.20\textwidth]{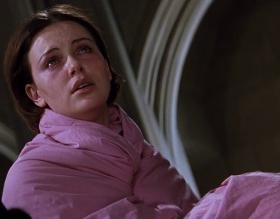}}
\end{tabular}
\end{center}
\end{table*}

\vspace{-1cm}

\begin{tcolorbox}[enhanced, breakable]
            \footnotesize
            \vspace{.5cm}
            \textbf{Human, MCC $1$:}\\

\textcolor{red}{[}Keanu\textcolor{red}{]}$^{\textcolor{red}{1}}$ tells \textcolor{blue}{[}Charlize\textcolor{blue}{]}$^{\textcolor{blue}{2}}$ how much \textcolor{red}{[}she\textcolor{red}{]}$^{\textcolor{red}{1}}$ loves \textcolor{blue}{[}her\textcolor{blue}{]}$^{\textcolor{blue}{2}}$ and asks for forgiveness .

\textcolor{blue}{[}Charlize\textcolor{blue}{]}$^{\textcolor{blue}{2}}$ says it is very difficult to forgive \textcolor{red}{[}him\textcolor{red}{]}$^{\textcolor{red}{1}}$ because \textcolor{red}{[}he\textcolor{red}{]}$^{\textcolor{red}{1}}$ hurt \textcolor{blue}{[}her\textcolor{blue}{]}$^{\textcolor{blue}{2}}$ so much .

\textcolor{blue}{[}Charlize\textcolor{blue}{]}$^{\textcolor{blue}{2}}$ cries a lot and \textcolor{red}{[}Keanu\textcolor{red}{]}$^{\textcolor{red}{1}}$ Loses forgiveness .

\textcolor{blue}{[}Charlize\textcolor{blue}{]}$^{\textcolor{blue}{2}}$ stops crying and says \textcolor{blue}{[}she\textcolor{blue}{]}$^{\textcolor{blue}{2}}$ loves \textcolor{red}{[}him\textcolor{red}{]}$^{\textcolor{red}{1}}$ so much .

who will trust \textcolor{red}{[}him\textcolor{red}{]}$^{\textcolor{red}{1}}$ one more time .

        \vspace{.5cm}
        \textbf{\modelname{DeepSeek-VL2-1B}, MCC $0.33$:}\\

\textcolor{red}{[}The sequence of images\textcolor{red}{]}$^{\textcolor{red}{1}}$ depicts \textcolor{orange}{[}a dramatic scene with \textcolor{blue}{[}two individuals\textcolor{blue}{]}$^{\textcolor{blue}{2}}$ engaged in an intense conversation\textcolor{orange}{]}$^{\textcolor{orange}{3}}$ .

The first image shows \textcolor{teal}{[}a man in a dark suit\textcolor{teal}{]}$^{\textcolor{teal}{4}}$ looking at \textcolor{purple}{[}a woman who appears distressed , \textcolor{brown}{[}\textcolor{purple}{[}her\textcolor{purple}{]}$^{\textcolor{purple}{5}}$ face\textcolor{brown}{]}$^{\textcolor{brown}{6}}$ contorted with emotion\textcolor{purple}{]}$^{\textcolor{purple}{5}}$ .

The second image captures \textcolor{purple}{[}the same woman\textcolor{purple}{]}$^{\textcolor{purple}{5}}$ , now wrapped in \textcolor{magenta}{[}a pink blanket\textcolor{magenta}{]}$^{\textcolor{magenta}{7}}$ , \textcolor{olive}{[}\textcolor{purple}{[}her\textcolor{purple}{]}$^{\textcolor{purple}{5}}$ eyes\textcolor{olive}{]}$^{\textcolor{olive}{8}}$ closed and \textcolor{cyan}{[}mouth\textcolor{cyan}{]}$^{\textcolor{cyan}{9}}$ open as if \textcolor{purple}{[}she\textcolor{purple}{]}$^{\textcolor{purple}{5}}$ is crying .

The third image shows \textcolor{purple}{[}the woman\textcolor{purple}{]}$^{\textcolor{purple}{5}}$ still wrapped in \textcolor{magenta}{[}the blanket\textcolor{magenta}{]}$^{\textcolor{magenta}{7}}$ , \textcolor{purple}{[}her\textcolor{purple}{]}$^{\textcolor{purple}{5}}$ expression more subdued , with \textcolor{purple}{[}her\textcolor{purple}{]}$^{\textcolor{purple}{5}}$ eyes partially closed .

The fourth image focuses on \textcolor{purple}{[}the woman 's\textcolor{purple}{]}$^{\textcolor{purple}{5}}$ face , showing \textcolor{purple}{[}her\textcolor{purple}{]}$^{\textcolor{purple}{5}}$ in a state of distress , with \textcolor{purple}{[}her\textcolor{purple}{]}$^{\textcolor{purple}{5}}$ eyes wide open and mouth slightly agape .

The fifth image is a close-up of \textcolor{purple}{[}the woman 's\textcolor{purple}{]}$^{\textcolor{purple}{5}}$ face , emphasizing \textcolor{purple}{[}her\textcolor{purple}{]}$^{\textcolor{purple}{5}}$ emotional state .

The story that fits \textcolor{red}{[}the sequence of images\textcolor{red}{]}$^{\textcolor{red}{1}}$ is as follows : In \textcolor{lime}{[}the dimly lit room\textcolor{lime}{]}$^{\textcolor{lime}{10}}$ , \textcolor{teal}{[}Keanu , a man in a dark suit\textcolor{teal}{]}$^{\textcolor{teal}{4}}$ , stands beside \textcolor{purple}{[}a woman wrapped in \textcolor{magenta}{[}a pink blanket\textcolor{magenta}{]}$^{\textcolor{magenta}{7}}$\textcolor{purple}{]}$^{\textcolor{purple}{5}}$ .

\textcolor{purple}{[}She\textcolor{purple}{]}$^{\textcolor{purple}{5}}$ appears to be in distress , \textcolor{brown}{[}\textcolor{purple}{[}her\textcolor{purple}{]}$^{\textcolor{purple}{5}}$ face\textcolor{brown}{]}$^{\textcolor{brown}{6}}$ contorted with emotion .

\textcolor{lime}{[}The room\textcolor{lime}{]}$^{\textcolor{lime}{10}}$ is dimly lit , with shadows cast on the walls , creating an atmosphere of tension .

\textcolor{teal}{[}Keanu\textcolor{teal}{]}$^{\textcolor{teal}{4}}$ looks at \textcolor{purple}{[}her\textcolor{purple}{]}$^{\textcolor{purple}{5}}$ with concern , \textcolor{teal}{[}his\textcolor{teal}{]}$^{\textcolor{teal}{4}}$ expression serious .

\textcolor{olive}{[}\textcolor{purple}{[}The woman 's\textcolor{purple}{]}$^{\textcolor{purple}{5}}$ eyes\textcolor{olive}{]}$^{\textcolor{olive}{8}}$ are closed , and \textcolor{cyan}{[}\textcolor{purple}{[}her\textcolor{purple}{]}$^{\textcolor{purple}{5}}$ mouth\textcolor{cyan}{]}$^{\textcolor{cyan}{9}}$ is open as if \textcolor{purple}{[}she\textcolor{purple}{]}$^{\textcolor{purple}{5}}$ is crying .

\textcolor{purple}{[}She\textcolor{purple}{]}$^{\textcolor{purple}{5}}$ seems to be in a state of emotional turmoil , possibly due to a recent event or revelation .

\textcolor{orange}{[}The scene\textcolor{orange}{]}$^{\textcolor{orange}{3}}$ suggests a moment of high tension , with \textcolor{blue}{[}the characters\textcolor{blue}{]}$^{\textcolor{blue}{2}}$ deeply engaged in a conversation that could have significant consequences .

        \vspace{.5cm}
        \textbf{\modelname{DeepSeek-VL2-4.5B}, MCC $0.68$:}\\
        
The story begins with \textcolor{red}{[}Keanu , a man in a black suit\textcolor{red}{]}$^{\textcolor{red}{1}}$ , standing close to \textcolor{blue}{[}Charlize , a woman wrapped in \textcolor{orange}{[}a pink blanket\textcolor{orange}{]}$^{\textcolor{orange}{3}}$\textcolor{blue}{]}$^{\textcolor{blue}{2}}$ , in \textcolor{teal}{[}a dimly lit room\textcolor{teal}{]}$^{\textcolor{teal}{4}}$ .

\textcolor{purple}{[}\textcolor{red}{[}Keanu 's\textcolor{red}{]}$^{\textcolor{red}{1}}$ expression\textcolor{purple}{]}$^{\textcolor{purple}{5}}$ is serious as \textcolor{red}{[}he\textcolor{red}{]}$^{\textcolor{red}{1}}$ leans in , \textcolor{red}{[}his\textcolor{red}{]}$^{\textcolor{red}{1}}$ eyes locked onto \textcolor{blue}{[}Charlize 's\textcolor{blue}{]}$^{\textcolor{blue}{2}}$ .

\textcolor{blue}{[}She\textcolor{blue}{]}$^{\textcolor{blue}{2}}$ looks up at \textcolor{red}{[}him\textcolor{red}{]}$^{\textcolor{red}{1}}$ , \textcolor{brown}{[}\textcolor{blue}{[}her\textcolor{blue}{]}$^{\textcolor{blue}{2}}$ face\textcolor{brown}{]}$^{\textcolor{brown}{6}}$ a mix of concern and confusion .

The tension between \textcolor{magenta}{[}them\textcolor{magenta}{]}$^{\textcolor{magenta}{7}}$ is palpable , as if \textcolor{magenta}{[}they\textcolor{magenta}{]}$^{\textcolor{magenta}{7}}$ are discussing something of great importance .

\textcolor{olive}{[}\textcolor{blue}{[}Charlize 's\textcolor{blue}{]}$^{\textcolor{blue}{2}}$ eyes\textcolor{olive}{]}$^{\textcolor{olive}{8}}$ well up with tears as \textcolor{blue}{[}she\textcolor{blue}{]}$^{\textcolor{blue}{2}}$ speaks , \textcolor{blue}{[}her\textcolor{blue}{]}$^{\textcolor{blue}{2}}$ voice trembling .

`` \textcolor{red}{[}Keanu\textcolor{red}{]}$^{\textcolor{red}{1}}$ , \textcolor{blue}{[}I\textcolor{blue}{]}$^{\textcolor{blue}{2}}$ ca n't keep doing \textcolor{cyan}{[}this\textcolor{cyan}{]}$^{\textcolor{cyan}{9}}$ , '' \textcolor{blue}{[}she\textcolor{blue}{]}$^{\textcolor{blue}{2}}$ says , \textcolor{blue}{[}her\textcolor{blue}{]}$^{\textcolor{blue}{2}}$ words barely audible .

`` \textcolor{blue}{[}I\textcolor{blue}{]}$^{\textcolor{blue}{2}}$ need to find a way out of \textcolor{cyan}{[}this\textcolor{cyan}{]}$^{\textcolor{cyan}{9}}$ . ''

\textcolor{purple}{[}\textcolor{red}{[}Keanu 's\textcolor{red}{]}$^{\textcolor{red}{1}}$ expression\textcolor{purple}{]}$^{\textcolor{purple}{5}}$ softens slightly , but \textcolor{red}{[}he\textcolor{red}{]}$^{\textcolor{red}{1}}$ remains firm .

`` \textcolor{blue}{[}You\textcolor{blue}{]}$^{\textcolor{blue}{2}}$ ca n't just leave , '' \textcolor{red}{[}he\textcolor{red}{]}$^{\textcolor{red}{1}}$ replies , \textcolor{red}{[}his\textcolor{red}{]}$^{\textcolor{red}{1}}$ tone gentle yet resolute .

As the conversation continues , \textcolor{blue}{[}Charlize 's\textcolor{blue}{]}$^{\textcolor{blue}{2}}$ distress becomes more evident .

\textcolor{blue}{[}She\textcolor{blue}{]}$^{\textcolor{blue}{2}}$ clutches \textcolor{orange}{[}the blanket\textcolor{orange}{]}$^{\textcolor{orange}{3}}$ around \textcolor{blue}{[}her\textcolor{blue}{]}$^{\textcolor{blue}{2}}$ , \textcolor{blue}{[}her\textcolor{blue}{]}$^{\textcolor{blue}{2}}$ body shaking with emotion .

\textcolor{red}{[}Keanu\textcolor{red}{]}$^{\textcolor{red}{1}}$ watches \textcolor{blue}{[}her\textcolor{blue}{]}$^{\textcolor{blue}{2}}$ , \textcolor{lime}{[}\textcolor{red}{[}his\textcolor{red}{]}$^{\textcolor{red}{1}}$ heart\textcolor{lime}{]}$^{\textcolor{lime}{10}}$ aching for \textcolor{blue}{[}her\textcolor{blue}{]}$^{\textcolor{blue}{2}}$ pain .

`` \textcolor{red}{[}I\textcolor{red}{]}$^{\textcolor{red}{1}}$ know it 's not easy , '' \textcolor{red}{[}he\textcolor{red}{]}$^{\textcolor{red}{1}}$ says , \textcolor{red}{[}his\textcolor{red}{]}$^{\textcolor{red}{1}}$ voice filled with empathy .

`` But \textcolor{magenta}{[}we\textcolor{magenta}{]}$^{\textcolor{magenta}{7}}$ have to find a way . ''

In the next image , \textcolor{olive}{[}\textcolor{blue}{[}Charlize 's\textcolor{blue}{]}$^{\textcolor{blue}{2}}$ eyes\textcolor{olive}{]}$^{\textcolor{olive}{8}}$ are closed , \textcolor{brown}{[}\textcolor{blue}{[}her\textcolor{blue}{]}$^{\textcolor{blue}{2}}$ face\textcolor{brown}{]}$^{\textcolor{brown}{6}}$ contorted in pain .

\textcolor{blue}{[}She\textcolor{blue}{]}$^{\textcolor{blue}{2}}$ takes a deep breath , trying to compose \textcolor{blue}{[}herself\textcolor{blue}{]}$^{\textcolor{blue}{2}}$ .

\textcolor{red}{[}Keanu\textcolor{red}{]}$^{\textcolor{red}{1}}$ stands beside \textcolor{blue}{[}her\textcolor{blue}{]}$^{\textcolor{blue}{2}}$ , \textcolor{red}{[}his\textcolor{red}{]}$^{\textcolor{red}{1}}$ hand gently resting on \textcolor{blue}{[}her\textcolor{blue}{]}$^{\textcolor{blue}{2}}$ shoulder .

`` \textcolor{blue}{[}You\textcolor{blue}{]}$^{\textcolor{blue}{2}}$ 're not alone , '' \textcolor{red}{[}he\textcolor{red}{]}$^{\textcolor{red}{1}}$ whispers , \textcolor{red}{[}his\textcolor{red}{]}$^{\textcolor{red}{1}}$ voice a soothing balm to \textcolor{blue}{[}her\textcolor{blue}{]}$^{\textcolor{blue}{2}}$ turmoil .

Finally , \textcolor{blue}{[}Charlize\textcolor{blue}{]}$^{\textcolor{blue}{2}}$ opens \textcolor{olive}{[}\textcolor{blue}{[}her\textcolor{blue}{]}$^{\textcolor{blue}{2}}$ eyes\textcolor{olive}{]}$^{\textcolor{olive}{8}}$ , \textcolor{blue}{[}her\textcolor{blue}{]}$^{\textcolor{blue}{2}}$ gaze filled with determination .

\textcolor{blue}{[}She\textcolor{blue}{]}$^{\textcolor{blue}{2}}$ reaches out with \textcolor{blue}{[}her\textcolor{blue}{]}$^{\textcolor{blue}{2}}$ hand , \textcolor{blue}{[}her\textcolor{blue}{]}$^{\textcolor{blue}{2}}$ fingers brushing against \textcolor{red}{[}Keanu 's\textcolor{red}{]}$^{\textcolor{red}{1}}$ .

`` \textcolor{magenta}{[}We\textcolor{magenta}{]}$^{\textcolor{magenta}{7}}$ 'll figure this out together , '' \textcolor{blue}{[}she\textcolor{blue}{]}$^{\textcolor{blue}{2}}$ says , \textcolor{blue}{[}her\textcolor{blue}{]}$^{\textcolor{blue}{2}}$ voice steady .

\textcolor{red}{[}Keanu\textcolor{red}{]}$^{\textcolor{red}{1}}$ smiles , a sense of hope filling \textcolor{lime}{[}\textcolor{red}{[}his\textcolor{red}{]}$^{\textcolor{red}{1}}$ heart\textcolor{lime}{]}$^{\textcolor{lime}{10}}$ .

`` Yes , '' \textcolor{red}{[}he\textcolor{red}{]}$^{\textcolor{red}{1}}$ replies , \textcolor{red}{[}his\textcolor{red}{]}$^{\textcolor{red}{1}}$ eyes shining with resolve .

As \textcolor{magenta}{[}they\textcolor{magenta}{]}$^{\textcolor{magenta}{7}}$ stand together , the dim light of \textcolor{teal}{[}the room\textcolor{teal}{]}$^{\textcolor{teal}{4}}$ casting a warm glow around \textcolor{magenta}{[}them\textcolor{magenta}{]}$^{\textcolor{magenta}{7}}$ , \textcolor{magenta}{[}they\textcolor{magenta}{]}$^{\textcolor{magenta}{7}}$ know that \textcolor{violet}{[}whatever challenges lie ahead\textcolor{violet}{]}$^{\textcolor{violet}{11}}$ , \textcolor{magenta}{[}they\textcolor{magenta}{]}$^{\textcolor{magenta}{7}}$ will face \textcolor{violet}{[}them\textcolor{violet}{]}$^{\textcolor{violet}{11}}$ as a team .

        \vspace{.5cm}
        \textbf{\modelname{Gemini 2.0 Flash}, MCC $0.66$:}\\

\textcolor{red}{[}Keanu\textcolor{red}{]}$^{\textcolor{red}{1}}$ looks intently at \textcolor{blue}{[}Charlize , who seems distressed with \textcolor{orange}{[}\textcolor{blue}{[}her\textcolor{blue}{]}$^{\textcolor{blue}{2}}$ hand\textcolor{orange}{]}$^{\textcolor{orange}{3}}$ to \textcolor{teal}{[}her face\textcolor{teal}{]}$^{\textcolor{teal}{4}}$\textcolor{blue}{]}$^{\textcolor{blue}{2}}$ .

\textcolor{blue}{[}Charlize\textcolor{blue}{]}$^{\textcolor{blue}{2}}$ appears more upset with tears streaming down \textcolor{teal}{[}\textcolor{blue}{[}her\textcolor{blue}{]}$^{\textcolor{blue}{2}}$ face\textcolor{teal}{]}$^{\textcolor{teal}{4}}$ and a blanket wrapped around \textcolor{blue}{[}her\textcolor{blue}{]}$^{\textcolor{blue}{2}}$ shoulders .

\textcolor{blue}{[}She\textcolor{blue}{]}$^{\textcolor{blue}{2}}$ looks up , perhaps pleading for answers .

In the third image , \textcolor{blue}{[}she\textcolor{blue}{]}$^{\textcolor{blue}{2}}$ closes \textcolor{blue}{[}her\textcolor{blue}{]}$^{\textcolor{blue}{2}}$ eyes , clearly in pain .

Then , in the next image \textcolor{blue}{[}she\textcolor{blue}{]}$^{\textcolor{blue}{2}}$ raises \textcolor{orange}{[}\textcolor{blue}{[}her\textcolor{blue}{]}$^{\textcolor{blue}{2}}$ hand\textcolor{orange}{]}$^{\textcolor{orange}{3}}$ as if to speak .

It seems a tragic event has left \textcolor{blue}{[}Charlize\textcolor{blue}{]}$^{\textcolor{blue}{2}}$ distraught , while \textcolor{red}{[}Keanu\textcolor{red}{]}$^{\textcolor{red}{1}}$ looks like \textcolor{red}{[}he\textcolor{red}{]}$^{\textcolor{red}{1}}$ is trying to help .

        \vspace{.5cm}
        \textbf{\modelname{GPT4o}, MCC $0.67$:}\\

\textcolor{red}{[}Keanu\textcolor{red}{]}$^{\textcolor{red}{1}}$ walked briskly through the echoing halls , urgency marking \textcolor{red}{[}his\textcolor{red}{]}$^{\textcolor{red}{1}}$ steps .

\textcolor{red}{[}He\textcolor{red}{]}$^{\textcolor{red}{1}}$ found \textcolor{blue}{[}Charlize\textcolor{blue}{]}$^{\textcolor{blue}{2}}$ wrapped in \textcolor{orange}{[}a pink blanket\textcolor{orange}{]}$^{\textcolor{orange}{3}}$ , tears streaming down \textcolor{blue}{[}her\textcolor{blue}{]}$^{\textcolor{blue}{2}}$ face .

\textcolor{teal}{[}\textcolor{blue}{[}Her\textcolor{blue}{]}$^{\textcolor{blue}{2}}$ eyes\textcolor{teal}{]}$^{\textcolor{teal}{4}}$ met \textcolor{red}{[}his\textcolor{red}{]}$^{\textcolor{red}{1}}$ , seeking comfort and understanding .

\textcolor{red}{[}Keanu\textcolor{red}{]}$^{\textcolor{red}{1}}$ kneeled beside \textcolor{blue}{[}her\textcolor{blue}{]}$^{\textcolor{blue}{2}}$ , concern etched on \textcolor{red}{[}his\textcolor{red}{]}$^{\textcolor{red}{1}}$ features .

\textcolor{blue}{[}Charlize\textcolor{blue}{]}$^{\textcolor{blue}{2}}$ looked up , \textcolor{blue}{[}her\textcolor{blue}{]}$^{\textcolor{blue}{2}}$ voice trembling as \textcolor{blue}{[}she\textcolor{blue}{]}$^{\textcolor{blue}{2}}$ tried to speak about what had shaken \textcolor{blue}{[}her\textcolor{blue}{]}$^{\textcolor{blue}{2}}$ so deeply .

\textcolor{blue}{[}Her\textcolor{blue}{]}$^{\textcolor{blue}{2}}$ vulnerability reached out to \textcolor{red}{[}him\textcolor{red}{]}$^{\textcolor{red}{1}}$ , and \textcolor{red}{[}he\textcolor{red}{]}$^{\textcolor{red}{1}}$ offered a steady hand , encouraging \textcolor{blue}{[}her\textcolor{blue}{]}$^{\textcolor{blue}{2}}$ to share \textcolor{blue}{[}her\textcolor{blue}{]}$^{\textcolor{blue}{2}}$ burden .

\textcolor{blue}{[}Charlize\textcolor{blue}{]}$^{\textcolor{blue}{2}}$ struggled , \textcolor{blue}{[}her\textcolor{blue}{]}$^{\textcolor{blue}{2}}$ words breaking under the weight of emotion .

But \textcolor{red}{[}Keanu\textcolor{red}{]}$^{\textcolor{red}{1}}$ remained patient , listening intently , \textcolor{red}{[}his\textcolor{red}{]}$^{\textcolor{red}{1}}$ presence a silent promise of support .

As \textcolor{blue}{[}she\textcolor{blue}{]}$^{\textcolor{blue}{2}}$ continued , \textcolor{blue}{[}her\textcolor{blue}{]}$^{\textcolor{blue}{2}}$ words began to flow more freely .

\textcolor{purple}{[}The fear that had gripped \textcolor{blue}{[}her\textcolor{blue}{]}$^{\textcolor{blue}{2}}$\textcolor{purple}{]}$^{\textcolor{purple}{5}}$ started to loosen \textcolor{purple}{[}its\textcolor{purple}{]}$^{\textcolor{purple}{5}}$ hold .

\textcolor{blue}{[}She\textcolor{blue}{]}$^{\textcolor{blue}{2}}$ clenched \textcolor{orange}{[}\textcolor{blue}{[}her\textcolor{blue}{]}$^{\textcolor{blue}{2}}$ blanket\textcolor{orange}{]}$^{\textcolor{orange}{3}}$ tighter , as if drawing strength from the warmth .

\textcolor{red}{[}Keanu ’ s\textcolor{red}{]}$^{\textcolor{red}{1}}$ soft gaze and reassuring nods communicated what words couldn't—she was not alone .

With each passing moment , the trust between \textcolor{brown}{[}them\textcolor{brown}{]}$^{\textcolor{brown}{6}}$ grew .

\textcolor{teal}{[}\textcolor{blue}{[}Charlize 's\textcolor{blue}{]}$^{\textcolor{blue}{2}}$ eyes\textcolor{teal}{]}$^{\textcolor{teal}{4}}$ , though red from tears , began to shine with a glimmer of relief as \textcolor{red}{[}Keanu\textcolor{red}{]}$^{\textcolor{red}{1}}$ reached out , bridging the space between \textcolor{brown}{[}them\textcolor{brown}{]}$^{\textcolor{brown}{6}}$ with understanding and care .

        \vspace{.5cm}
        \textbf{\modelname{InternVL2.5-78B}, MCC $0.72$:}\\

\textcolor{orange}{[}\textcolor{red}{[}Keanu\textcolor{red}{]}$^{\textcolor{red}{1}}$ and \textcolor{blue}{[}Charlize\textcolor{blue}{]}$^{\textcolor{blue}{2}}$\textcolor{orange}{]}$^{\textcolor{orange}{3}}$ find \textcolor{orange}{[}themselves\textcolor{orange}{]}$^{\textcolor{orange}{3}}$ alone in a dimly lit room , \textcolor{teal}{[}\textcolor{orange}{[}their\textcolor{orange}{]}$^{\textcolor{orange}{3}}$ conversation\textcolor{teal}{]}$^{\textcolor{teal}{4}}$ marked by tension and emotion .

\textcolor{red}{[}Keanu\textcolor{red}{]}$^{\textcolor{red}{1}}$ , dressed in a formal black suit , seems desperate to explain \textcolor{red}{[}himself\textcolor{red}{]}$^{\textcolor{red}{1}}$ .

\textcolor{red}{[}He\textcolor{red}{]}$^{\textcolor{red}{1}}$ leans forward , looking serious but hopeful as \textcolor{red}{[}he\textcolor{red}{]}$^{\textcolor{red}{1}}$ speaks to \textcolor{blue}{[}Charlize\textcolor{blue}{]}$^{\textcolor{blue}{2}}$ .

The flickering light behind \textcolor{red}{[}him\textcolor{red}{]}$^{\textcolor{red}{1}}$ casts a somber glow , adding to the gravity of \textcolor{teal}{[}the conversation\textcolor{teal}{]}$^{\textcolor{teal}{4}}$ .

\textcolor{blue}{[}Charlize\textcolor{blue}{]}$^{\textcolor{blue}{2}}$ , wrapped in \textcolor{purple}{[}a pink blanket\textcolor{purple}{]}$^{\textcolor{purple}{5}}$ , sits tensely .

\textcolor{brown}{[}\textcolor{blue}{[}Her\textcolor{blue}{]}$^{\textcolor{blue}{2}}$ eyes\textcolor{brown}{]}$^{\textcolor{brown}{6}}$ are filled with a mix of sadness and confusion as \textcolor{blue}{[}she\textcolor{blue}{]}$^{\textcolor{blue}{2}}$ listens to \textcolor{red}{[}Keanu\textcolor{red}{]}$^{\textcolor{red}{1}}$ .

\textcolor{blue}{[}She\textcolor{blue}{]}$^{\textcolor{blue}{2}}$ touches \textcolor{blue}{[}her\textcolor{blue}{]}$^{\textcolor{blue}{2}}$ head , \textcolor{blue}{[}her\textcolor{blue}{]}$^{\textcolor{blue}{2}}$ hand trembling slightly .

Tears well up in \textcolor{brown}{[}\textcolor{blue}{[}her\textcolor{blue}{]}$^{\textcolor{blue}{2}}$ eyes\textcolor{brown}{]}$^{\textcolor{brown}{6}}$ as \textcolor{blue}{[}she\textcolor{blue}{]}$^{\textcolor{blue}{2}}$ processes \textcolor{red}{[}his\textcolor{red}{]}$^{\textcolor{red}{1}}$ words , \textcolor{blue}{[}her\textcolor{blue}{]}$^{\textcolor{blue}{2}}$ gaze occasionally flickering away from \textcolor{red}{[}him\textcolor{red}{]}$^{\textcolor{red}{1}}$ .

The atmosphere thickens with emotion .

\textcolor{red}{[}Keanu ’ s\textcolor{red}{]}$^{\textcolor{red}{1}}$ expression shifts from hopeful to anxious as the significance of what \textcolor{red}{[}he\textcolor{red}{]}$^{\textcolor{red}{1}}$ is saying becomes more intense .

\textcolor{red}{[}He\textcolor{red}{]}$^{\textcolor{red}{1}}$ reaches out , as if to bridge the emotional gap between \textcolor{orange}{[}them\textcolor{orange}{]}$^{\textcolor{orange}{3}}$ , but pulls back , realizing the depth of \textcolor{blue}{[}Charlize ’ s\textcolor{blue}{]}$^{\textcolor{blue}{2}}$ turmoil .

\textcolor{blue}{[}Charlize\textcolor{blue}{]}$^{\textcolor{blue}{2}}$ , clearly overwhelmed , looks down , \textcolor{blue}{[}her\textcolor{blue}{]}$^{\textcolor{blue}{2}}$ face reflecting the inner battle \textcolor{blue}{[}she\textcolor{blue}{]}$^{\textcolor{blue}{2}}$ is fighting .

The soft pink of \textcolor{purple}{[}\textcolor{blue}{[}her\textcolor{blue}{]}$^{\textcolor{blue}{2}}$ blanket\textcolor{purple}{]}$^{\textcolor{purple}{5}}$ contrasts sharply with the seriousness of \textcolor{blue}{[}her\textcolor{blue}{]}$^{\textcolor{blue}{2}}$ emotions .

\textcolor{red}{[}Keanu\textcolor{red}{]}$^{\textcolor{red}{1}}$ , sensing \textcolor{blue}{[}her\textcolor{blue}{]}$^{\textcolor{blue}{2}}$ distress , attempts to comfort \textcolor{blue}{[}her\textcolor{blue}{]}$^{\textcolor{blue}{2}}$ , \textcolor{red}{[}his\textcolor{red}{]}$^{\textcolor{red}{1}}$ voice taking on a more soothing tone .

Finally , \textcolor{blue}{[}Charlize\textcolor{blue}{]}$^{\textcolor{blue}{2}}$ raises \textcolor{brown}{[}\textcolor{blue}{[}her\textcolor{blue}{]}$^{\textcolor{blue}{2}}$ tear-filled eyes\textcolor{brown}{]}$^{\textcolor{brown}{6}}$ , a mix of sadness and acceptance visible in \textcolor{blue}{[}her\textcolor{blue}{]}$^{\textcolor{blue}{2}}$ gaze .

\textcolor{red}{[}Keanu\textcolor{red}{]}$^{\textcolor{red}{1}}$ looks back at \textcolor{blue}{[}her\textcolor{blue}{]}$^{\textcolor{blue}{2}}$ , \textcolor{red}{[}his\textcolor{red}{]}$^{\textcolor{red}{1}}$ expression softening into relief and love .

The image sequence ends with a tender moment , as \textcolor{blue}{[}Charlize\textcolor{blue}{]}$^{\textcolor{blue}{2}}$ , despite \textcolor{blue}{[}her\textcolor{blue}{]}$^{\textcolor{blue}{2}}$ sorrow , gives a faint nod , signaling that \textcolor{blue}{[}she\textcolor{blue}{]}$^{\textcolor{blue}{2}}$ understands and forgives \textcolor{red}{[}him\textcolor{red}{]}$^{\textcolor{red}{1}}$ .

The tension between \textcolor{orange}{[}them\textcolor{orange}{]}$^{\textcolor{orange}{3}}$ eases , replaced by a poignant understanding .

        \vspace{.5cm}
        \textbf{\modelname{Qwen2-VL-72B}, MCC 0.76}\\

\textcolor{orange}{[}\textcolor{red}{[}Keanu\textcolor{red}{]}$^{\textcolor{red}{1}}$ and \textcolor{blue}{[}Charlize\textcolor{blue}{]}$^{\textcolor{blue}{2}}$\textcolor{orange}{]}$^{\textcolor{orange}{3}}$ found \textcolor{orange}{[}themselves\textcolor{orange}{]}$^{\textcolor{orange}{3}}$ in a mysterious , dimly lit room with high arches and intricate designs on the walls .

\textcolor{orange}{[}They\textcolor{orange}{]}$^{\textcolor{orange}{3}}$ had stumbled upon \textcolor{teal}{[}an ancient artifact\textcolor{teal}{]}$^{\textcolor{teal}{4}}$ during \textcolor{orange}{[}their\textcolor{orange}{]}$^{\textcolor{orange}{3}}$ travels , and \textcolor{orange}{[}they\textcolor{orange}{]}$^{\textcolor{orange}{3}}$ were both captivated by \textcolor{purple}{[}\textcolor{teal}{[}its\textcolor{teal}{]}$^{\textcolor{teal}{4}}$ power\textcolor{purple}{]}$^{\textcolor{purple}{5}}$ .

As \textcolor{orange}{[}they\textcolor{orange}{]}$^{\textcolor{orange}{3}}$ examined \textcolor{teal}{[}the artifact\textcolor{teal}{]}$^{\textcolor{teal}{4}}$ , \textcolor{blue}{[}Charlize\textcolor{blue}{]}$^{\textcolor{blue}{2}}$ suddenly felt a sharp pain in \textcolor{blue}{[}her\textcolor{blue}{]}$^{\textcolor{blue}{2}}$ chest .

\textcolor{blue}{[}She\textcolor{blue}{]}$^{\textcolor{blue}{2}}$ clutched \textcolor{blue}{[}her\textcolor{blue}{]}$^{\textcolor{blue}{2}}$ heart and staggered backward , \textcolor{brown}{[}\textcolor{blue}{[}her\textcolor{blue}{]}$^{\textcolor{blue}{2}}$ face\textcolor{brown}{]}$^{\textcolor{brown}{6}}$ contorted in agony .

\textcolor{red}{[}Keanu\textcolor{red}{]}$^{\textcolor{red}{1}}$ rushed to \textcolor{blue}{[}her\textcolor{blue}{]}$^{\textcolor{blue}{2}}$ side , \textcolor{magenta}{[}\textcolor{red}{[}his\textcolor{red}{]}$^{\textcolor{red}{1}}$ eyes\textcolor{magenta}{]}$^{\textcolor{magenta}{7}}$ filled with concern .

\textcolor{red}{[}He\textcolor{red}{]}$^{\textcolor{red}{1}}$ tried to help \textcolor{blue}{[}her\textcolor{blue}{]}$^{\textcolor{blue}{2}}$ up , but \textcolor{blue}{[}she\textcolor{blue}{]}$^{\textcolor{blue}{2}}$ was too weak to stand .

\textcolor{olive}{[}\textcolor{blue}{[}Charlize 's\textcolor{blue}{]}$^{\textcolor{blue}{2}}$ eyes\textcolor{olive}{]}$^{\textcolor{olive}{8}}$ rolled back , and \textcolor{blue}{[}she\textcolor{blue}{]}$^{\textcolor{blue}{2}}$ collapsed onto the floor .

\textcolor{red}{[}Keanu\textcolor{red}{]}$^{\textcolor{red}{1}}$ knelt beside \textcolor{blue}{[}her\textcolor{blue}{]}$^{\textcolor{blue}{2}}$ , gently cradling \textcolor{blue}{[}her\textcolor{blue}{]}$^{\textcolor{blue}{2}}$ head in \textcolor{red}{[}his\textcolor{red}{]}$^{\textcolor{red}{1}}$ hands .

\textcolor{red}{[}He\textcolor{red}{]}$^{\textcolor{red}{1}}$ could see the fear and confusion in \textcolor{olive}{[}\textcolor{blue}{[}her\textcolor{blue}{]}$^{\textcolor{blue}{2}}$ eyes\textcolor{olive}{]}$^{\textcolor{olive}{8}}$ as \textcolor{blue}{[}she\textcolor{blue}{]}$^{\textcolor{blue}{2}}$ struggled to breathe .

With a deep breath , \textcolor{blue}{[}Charlize\textcolor{blue}{]}$^{\textcolor{blue}{2}}$ managed to speak .

`` What ... what 's happening to \textcolor{blue}{[}me\textcolor{blue}{]}$^{\textcolor{blue}{2}}$ ? ''

\textcolor{blue}{[}she\textcolor{blue}{]}$^{\textcolor{blue}{2}}$ gasped .

\textcolor{red}{[}Keanu\textcolor{red}{]}$^{\textcolor{red}{1}}$ looked around frantically , searching for any clue that might explain \textcolor{blue}{[}her\textcolor{blue}{]}$^{\textcolor{blue}{2}}$ sudden collapse .

Suddenly , \textcolor{red}{[}he\textcolor{red}{]}$^{\textcolor{red}{1}}$ noticed \textcolor{cyan}{[}a faint glow\textcolor{cyan}{]}$^{\textcolor{cyan}{9}}$ emanating from \textcolor{teal}{[}the artifact\textcolor{teal}{]}$^{\textcolor{teal}{4}}$ .

As \textcolor{red}{[}he\textcolor{red}{]}$^{\textcolor{red}{1}}$ reached out to touch \textcolor{teal}{[}it\textcolor{teal}{]}$^{\textcolor{teal}{4}}$ , \textcolor{red}{[}he\textcolor{red}{]}$^{\textcolor{red}{1}}$ felt a surge of energy coursing through \textcolor{red}{[}his\textcolor{red}{]}$^{\textcolor{red}{1}}$ body .

\textcolor{red}{[}He\textcolor{red}{]}$^{\textcolor{red}{1}}$ realized that \textcolor{teal}{[}the artifact\textcolor{teal}{]}$^{\textcolor{teal}{4}}$ was somehow linked to \textcolor{blue}{[}Charlize 's\textcolor{blue}{]}$^{\textcolor{blue}{2}}$ condition .

\textcolor{red}{[}Keanu\textcolor{red}{]}$^{\textcolor{red}{1}}$ focused all \textcolor{red}{[}his\textcolor{red}{]}$^{\textcolor{red}{1}}$ energy on \textcolor{teal}{[}the artifact\textcolor{teal}{]}$^{\textcolor{teal}{4}}$ , willing \textcolor{teal}{[}it\textcolor{teal}{]}$^{\textcolor{teal}{4}}$ to reveal \textcolor{teal}{[}its\textcolor{teal}{]}$^{\textcolor{teal}{4}}$ secrets .

Slowly , \textcolor{cyan}{[}the glow\textcolor{cyan}{]}$^{\textcolor{cyan}{9}}$ intensified , and \textcolor{lime}{[}a shimmering image\textcolor{lime}{]}$^{\textcolor{lime}{10}}$ appeared before \textcolor{red}{[}him\textcolor{red}{]}$^{\textcolor{red}{1}}$ .

\textcolor{lime}{[}It\textcolor{lime}{]}$^{\textcolor{lime}{10}}$ showed \textcolor{blue}{[}Charlize\textcolor{blue}{]}$^{\textcolor{blue}{2}}$ , healthy and vibrant , standing in front of \textcolor{teal}{[}the same artifact\textcolor{teal}{]}$^{\textcolor{teal}{4}}$ .

\textcolor{red}{[}Keanu\textcolor{red}{]}$^{\textcolor{red}{1}}$ knew what \textcolor{red}{[}he\textcolor{red}{]}$^{\textcolor{red}{1}}$ had to do .

\textcolor{red}{[}He\textcolor{red}{]}$^{\textcolor{red}{1}}$ closed \textcolor{magenta}{[}\textcolor{red}{[}his\textcolor{red}{]}$^{\textcolor{red}{1}}$ eyes\textcolor{magenta}{]}$^{\textcolor{magenta}{7}}$ and concentrated , channeling \textcolor{purple}{[}\textcolor{teal}{[}the artifact 's\textcolor{teal}{]}$^{\textcolor{teal}{4}}$ power\textcolor{purple}{]}$^{\textcolor{purple}{5}}$ into \textcolor{blue}{[}Charlize 's\textcolor{blue}{]}$^{\textcolor{blue}{2}}$ body .

Slowly , \textcolor{blue}{[}her\textcolor{blue}{]}$^{\textcolor{blue}{2}}$ breathing steadied , and \textcolor{olive}{[}\textcolor{blue}{[}her\textcolor{blue}{]}$^{\textcolor{blue}{2}}$ eyes\textcolor{olive}{]}$^{\textcolor{olive}{8}}$ fluttered open .

\textcolor{blue}{[}She\textcolor{blue}{]}$^{\textcolor{blue}{2}}$ looked up at \textcolor{red}{[}Keanu\textcolor{red}{]}$^{\textcolor{red}{1}}$ , a smile spreading across \textcolor{brown}{[}\textcolor{blue}{[}her\textcolor{blue}{]}$^{\textcolor{blue}{2}}$ face\textcolor{brown}{]}$^{\textcolor{brown}{6}}$ .

`` Thank \textcolor{red}{[}you\textcolor{red}{]}$^{\textcolor{red}{1}}$ , '' \textcolor{blue}{[}she\textcolor{blue}{]}$^{\textcolor{blue}{2}}$ whispered .

\textcolor{red}{[}Keanu\textcolor{red}{]}$^{\textcolor{red}{1}}$ smiled back , relieved that \textcolor{blue}{[}she\textcolor{blue}{]}$^{\textcolor{blue}{2}}$ was safe .

Together , \textcolor{orange}{[}they\textcolor{orange}{]}$^{\textcolor{orange}{3}}$ stood up , ready to continue \textcolor{orange}{[}their\textcolor{orange}{]}$^{\textcolor{orange}{3}}$ journey , knowing that \textcolor{orange}{[}they\textcolor{orange}{]}$^{\textcolor{orange}{3}}$ had uncovered something truly extraordinary .

        \vspace{.5cm}
        \textbf{\modelname{Qwen2-VL-7B}, MCC $0.83$:}\\

In the quiet of \textcolor{red}{[}the room\textcolor{red}{]}$^{\textcolor{red}{1}}$ , \textcolor{teal}{[}\textcolor{blue}{[}Keanu\textcolor{blue}{]}$^{\textcolor{blue}{2}}$ and \textcolor{orange}{[}Charlize\textcolor{orange}{]}$^{\textcolor{orange}{3}}$\textcolor{teal}{]}$^{\textcolor{teal}{4}}$ stood facing each other , \textcolor{teal}{[}their\textcolor{teal}{]}$^{\textcolor{teal}{4}}$ eyes locked in a tense conversation .

\textcolor{orange}{[}Charlize 's\textcolor{orange}{]}$^{\textcolor{orange}{3}}$ face was marked with \textcolor{purple}{[}tears\textcolor{purple}{]}$^{\textcolor{purple}{5}}$ , and \textcolor{brown}{[}\textcolor{orange}{[}her\textcolor{orange}{]}$^{\textcolor{orange}{3}}$ body language\textcolor{brown}{]}$^{\textcolor{brown}{6}}$ suggested \textcolor{orange}{[}she\textcolor{orange}{]}$^{\textcolor{orange}{3}}$ was in distress .

\textcolor{blue}{[}Keanu\textcolor{blue}{]}$^{\textcolor{blue}{2}}$ , with a concerned expression , seemed to be trying to comfort \textcolor{orange}{[}her\textcolor{orange}{]}$^{\textcolor{orange}{3}}$ .

As the scene progressed , \textcolor{purple}{[}\textcolor{orange}{[}Charlize 's\textcolor{orange}{]}$^{\textcolor{orange}{3}}$ tears\textcolor{purple}{]}$^{\textcolor{purple}{5}}$ flowed freely , and \textcolor{brown}{[}\textcolor{orange}{[}her\textcolor{orange}{]}$^{\textcolor{orange}{3}}$ body language\textcolor{brown}{]}$^{\textcolor{brown}{6}}$ became more vulnerable .

\textcolor{blue}{[}Keanu\textcolor{blue}{]}$^{\textcolor{blue}{2}}$ , still looking at \textcolor{orange}{[}her\textcolor{orange}{]}$^{\textcolor{orange}{3}}$ with a mixture of concern and empathy , seemed to be trying to understand \textcolor{orange}{[}her\textcolor{orange}{]}$^{\textcolor{orange}{3}}$ pain .

\textcolor{red}{[}The room\textcolor{red}{]}$^{\textcolor{red}{1}}$ was filled with an intense atmosphere , and the tension between \textcolor{teal}{[}the two characters\textcolor{teal}{]}$^{\textcolor{teal}{4}}$ was palpable .
\end{tcolorbox}

%\iffalse

%\clearpage

\noindent
\begin{table*}[th]
\section{Example III}
\label{sec:examples3}

\iffalse
now, very low multimodal human continuity

% 2057
deepseek big 0.80
deepseek small 0.90
human 0.4
intern vl 0.60
qwen72 0.61
qwen small nan
gemini 0.57
gpt 0.80
\fi

\vspace{.5cm}

\begin{tabular}{cc}
  \shortstack{\textbf{Image 1}\\ \includegraphics[width=0.48\textwidth]{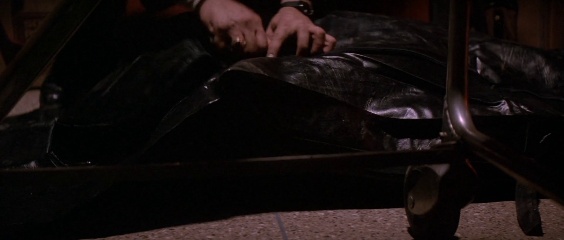}} &
  \shortstack{\textbf{Image 2}\\ \includegraphics[width=0.48\textwidth]{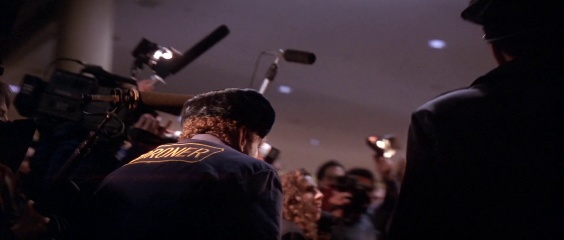}} \\[6pt]
  \shortstack{\textbf{Image 3}\\ \includegraphics[width=0.48\textwidth]{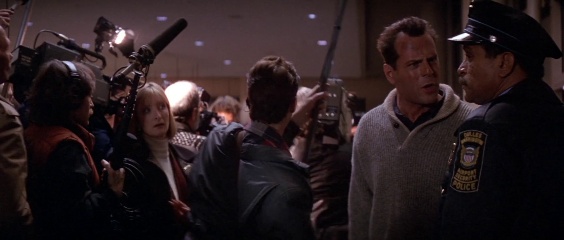}} &
  \shortstack{\textbf{Image 4}\\ \includegraphics[width=0.48\textwidth]{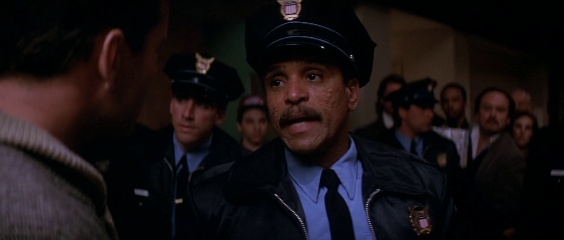}} \\[6pt]
  \multicolumn{2}{c}{%
    \shortstack{\textbf{Image 5}\\ \includegraphics[width=0.48\textwidth]{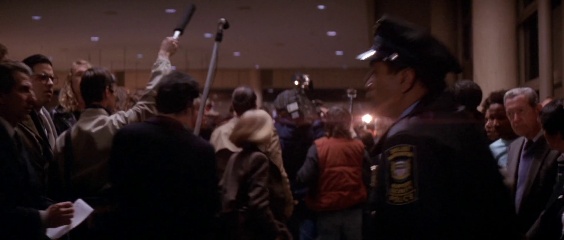}}
  }\\
\end{tabular}

\begin{center}
\begin{tabular}{ccc}
\centering
  \shortstack{\textbf{Character image 1: Bruce}\\ \includegraphics[width=0.20\textwidth]{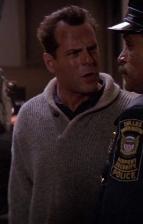}} 
\end{tabular}
\end{center}

\end{table*}

\vspace{-1cm}

\begin{tcolorbox}[enhanced, breakable]
            \footnotesize
            \vspace{.5cm}
            \textbf{Human:}\\

\textcolor{red}{[}The police officer\textcolor{red}{]}$^{\textcolor{red}{1}}$ closed the zipper of the body bag .

\textcolor{blue}{[}Journalists\textcolor{blue}{]}$^{\textcolor{blue}{2}}$ were already flooding the place .

\textcolor{orange}{[}Bruce\textcolor{orange}{]}$^{\textcolor{orange}{3}}$ asked \textcolor{red}{[}the officer\textcolor{red}{]}$^{\textcolor{red}{1}}$ how the hell \textcolor{blue}{[}all these journalists\textcolor{blue}{]}$^{\textcolor{blue}{2}}$ know already about the dead body .

\textcolor{red}{[}The police officer\textcolor{red}{]}$^{\textcolor{red}{1}}$ looked desperate .

\textcolor{red}{[}He\textcolor{red}{]}$^{\textcolor{red}{1}}$ said that every time had been like that in the last months .

They mast have a snitch in the station .

\textcolor{red}{[}He\textcolor{red}{]}$^{\textcolor{red}{1}}$ left \textcolor{orange}{[}Bruce\textcolor{orange}{]}$^{\textcolor{orange}{3}}$ and headed to \textcolor{blue}{[}the crowd of journalists\textcolor{blue}{]}$^{\textcolor{blue}{2}}$ .

It was about time for some order there .

        \vspace{.5cm}
        \textbf{\modelname{DeepSeek-VL2-1B}}\\

\textcolor{red}{[}The sequence of images\textcolor{red}{]}$^{\textcolor{red}{1}}$ depicts a chaotic scene with multiple individuals involved in what appears to be a public disturbance or protest .

\textcolor{blue}{[}The central figure , Bruce\textcolor{blue}{]}$^{\textcolor{blue}{2}}$ , is a police officer who seems to be addressing \textcolor{orange}{[}the crowd\textcolor{orange}{]}$^{\textcolor{orange}{3}}$ .

\textcolor{blue}{[}He\textcolor{blue}{]}$^{\textcolor{blue}{2}}$ is surrounded by other officers and a group of people , some of whom are holding cameras and microphones , indicating media presence .

The atmosphere is tense , with \textcolor{orange}{[}the crowd\textcolor{orange}{]}$^{\textcolor{orange}{3}}$ appearing agitated and \textcolor{blue}{[}the officer 's\textcolor{blue}{]}$^{\textcolor{blue}{2}}$ expression serious .

The story that fits \textcolor{red}{[}the sequence of images\textcolor{red}{]}$^{\textcolor{red}{1}}$ is as follows : \textcolor{blue}{[}Bruce , a dedicated police officer\textcolor{blue}{]}$^{\textcolor{blue}{2}}$ , is assigned to a high-profile case involving \textcolor{teal}{[}a viral social media video that has captured the attention of \textcolor{purple}{[}the public\textcolor{purple}{]}$^{\textcolor{purple}{5}}$\textcolor{teal}{]}$^{\textcolor{teal}{4}}$ .

\textcolor{teal}{[}The video , which depicts a young woman being attacked\textcolor{teal}{]}$^{\textcolor{teal}{4}}$ , has gone viral , sparking widespread outrage and \textcolor{brown}{[}protests\textcolor{brown}{]}$^{\textcolor{brown}{6}}$ .

As \textcolor{blue}{[}the officer\textcolor{blue}{]}$^{\textcolor{blue}{2}}$ , \textcolor{blue}{[}Bruce\textcolor{blue}{]}$^{\textcolor{blue}{2}}$ is tasked with maintaining order and ensuring the safety of the protesters while also handling \textcolor{magenta}{[}the media 's\textcolor{magenta}{]}$^{\textcolor{magenta}{7}}$ demands for information .

As the situation escalates , \textcolor{blue}{[}Bruce\textcolor{blue}{]}$^{\textcolor{blue}{2}}$ finds \textcolor{blue}{[}himself\textcolor{blue}{]}$^{\textcolor{blue}{2}}$ at the center of the conflict , facing pressure from both \textcolor{purple}{[}the public\textcolor{purple}{]}$^{\textcolor{purple}{5}}$ and \textcolor{magenta}{[}the media\textcolor{magenta}{]}$^{\textcolor{magenta}{7}}$ .

\textcolor{blue}{[}He\textcolor{blue}{]}$^{\textcolor{blue}{2}}$ must navigate the complexities of balancing \textcolor{blue}{[}his\textcolor{blue}{]}$^{\textcolor{blue}{2}}$ duty to protect \textcolor{olive}{[}the community\textcolor{olive}{]}$^{\textcolor{olive}{8}}$ with the demands of \textcolor{magenta}{[}the press\textcolor{magenta}{]}$^{\textcolor{magenta}{7}}$ .

Despite the challenges , \textcolor{blue}{[}Bruce\textcolor{blue}{]}$^{\textcolor{blue}{2}}$ remains committed to \textcolor{cyan}{[}doing\textcolor{cyan}{]}$^{\textcolor{cyan}{9}}$ what is right , even if \textcolor{cyan}{[}it\textcolor{cyan}{]}$^{\textcolor{cyan}{9}}$ means facing criticism and scrutiny .

In the end , \textcolor{blue}{[}Bruce 's\textcolor{blue}{]}$^{\textcolor{blue}{2}}$ actions lead to a resolution that addresses the underlying issues of \textcolor{teal}{[}the video\textcolor{teal}{]}$^{\textcolor{teal}{4}}$ and \textcolor{brown}{[}the protests\textcolor{brown}{]}$^{\textcolor{brown}{6}}$ , ultimately bringing closure to \textcolor{olive}{[}the community\textcolor{olive}{]}$^{\textcolor{olive}{8}}$ while also highlighting the importance of responsible journalism and public service .

        \vspace{.5cm}
        \textbf{\modelname{DeepSeek-VL2-4.5B}}\\

\textcolor{red}{[}Bruce\textcolor{red}{]}$^{\textcolor{red}{1}}$ stood in \textcolor{blue}{[}the dimly lit room\textcolor{blue}{]}$^{\textcolor{blue}{2}}$ , \textcolor{red}{[}his\textcolor{red}{]}$^{\textcolor{red}{1}}$ hands trembling as \textcolor{red}{[}he\textcolor{red}{]}$^{\textcolor{red}{1}}$ clutched \textcolor{orange}{[}the black plastic bag\textcolor{orange}{]}$^{\textcolor{orange}{3}}$ .

\textcolor{teal}{[}The weight of the unknown inside\textcolor{teal}{]}$^{\textcolor{teal}{4}}$ pressed down on \textcolor{red}{[}him\textcolor{red}{]}$^{\textcolor{red}{1}}$ , a heavy burden \textcolor{red}{[}he\textcolor{red}{]}$^{\textcolor{red}{1}}$ had to carry .

\textcolor{brown}{[}The faint blue lines on \textcolor{purple}{[}the floor\textcolor{purple}{]}$^{\textcolor{purple}{5}}$\textcolor{brown}{]}$^{\textcolor{brown}{6}}$ seemed to mock \textcolor{red}{[}him\textcolor{red}{]}$^{\textcolor{red}{1}}$ , a reminder of the mundane world \textcolor{red}{[}he\textcolor{red}{]}$^{\textcolor{red}{1}}$ was leaving behind .

\textcolor{red}{[}He\textcolor{red}{]}$^{\textcolor{red}{1}}$ glanced at \textcolor{magenta}{[}the metal cart\textcolor{magenta}{]}$^{\textcolor{magenta}{7}}$ , \textcolor{magenta}{[}its\textcolor{magenta}{]}$^{\textcolor{magenta}{7}}$ wheels cold and unyielding , a symbol of the journey ahead .

The shadows danced around \textcolor{red}{[}him\textcolor{red}{]}$^{\textcolor{red}{1}}$ , whispering secrets \textcolor{red}{[}he\textcolor{red}{]}$^{\textcolor{red}{1}}$ could n't yet decipher .

As \textcolor{red}{[}he\textcolor{red}{]}$^{\textcolor{red}{1}}$ stepped into \textcolor{olive}{[}the bustling crowd\textcolor{olive}{]}$^{\textcolor{olive}{8}}$ , \textcolor{teal}{[}the weight of \textcolor{orange}{[}the bag\textcolor{orange}{]}$^{\textcolor{orange}{3}}$\textcolor{teal}{]}$^{\textcolor{teal}{4}}$ felt even heavier .

\textcolor{cyan}{[}The cameras\textcolor{cyan}{]}$^{\textcolor{cyan}{9}}$ flashed , \textcolor{cyan}{[}their\textcolor{cyan}{]}$^{\textcolor{cyan}{9}}$ lights blinding \textcolor{red}{[}him\textcolor{red}{]}$^{\textcolor{red}{1}}$ momentarily .

\textcolor{red}{[}He\textcolor{red}{]}$^{\textcolor{red}{1}}$ saw \textcolor{lime}{[}the man in the blue jacket , the word `` POLICE '' emblazoned on \textcolor{lime}{[}his\textcolor{lime}{]}$^{\textcolor{lime}{10}}$ back\textcolor{lime}{]}$^{\textcolor{lime}{10}}$ , and \textcolor{violet}{[}\textcolor{olive}{[}the crowd 's\textcolor{olive}{]}$^{\textcolor{olive}{8}}$ murmurs\textcolor{violet}{]}$^{\textcolor{violet}{11}}$ grew louder .

\textcolor{red}{[}Bruce 's\textcolor{red}{]}$^{\textcolor{red}{1}}$ heart raced , \textcolor{red}{[}his\textcolor{red}{]}$^{\textcolor{red}{1}}$ mind a whirlwind of questions .

Who was \textcolor{lime}{[}he\textcolor{lime}{]}$^{\textcolor{lime}{10}}$ ?

What had \textcolor{lime}{[}he\textcolor{lime}{]}$^{\textcolor{lime}{10}}$ done ?

\textcolor{lime}{[}The police officer 's\textcolor{lime}{]}$^{\textcolor{lime}{10}}$ stern gaze met \textcolor{red}{[}his\textcolor{red}{]}$^{\textcolor{red}{1}}$ , and for a moment , time stood still .

\textcolor{red}{[}Bruce 's\textcolor{red}{]}$^{\textcolor{red}{1}}$ voice trembled as \textcolor{red}{[}he\textcolor{red}{]}$^{\textcolor{red}{1}}$ spoke , `` \textcolor{red}{[}I\textcolor{red}{]}$^{\textcolor{red}{1}}$ ...

\textcolor{red}{[}I\textcolor{red}{]}$^{\textcolor{red}{1}}$ do n't know what 's inside . ''

\textcolor{lime}{[}The officer 's\textcolor{lime}{]}$^{\textcolor{lime}{10}}$ expression softened , but \textcolor{violet}{[}\textcolor{olive}{[}the crowd 's\textcolor{olive}{]}$^{\textcolor{olive}{8}}$ murmurs\textcolor{violet}{]}$^{\textcolor{violet}{11}}$ grew louder .

\textcolor{red}{[}The man in the gray sweater\textcolor{red}{]}$^{\textcolor{red}{12}}$ stepped forward , \textcolor{red}{[}his\textcolor{red}{]}$^{\textcolor{red}{12}}$ eyes filled with concern .

`` We 'll help \textcolor{red}{[}you\textcolor{red}{]}$^{\textcolor{red}{1}}$ , '' \textcolor{red}{[}he\textcolor{red}{]}$^{\textcolor{red}{12}}$ said , \textcolor{red}{[}his\textcolor{red}{]}$^{\textcolor{red}{12}}$ voice steady .

\textcolor{red}{[}Bruce\textcolor{red}{]}$^{\textcolor{red}{1}}$ nodded , a small smile tugging at \textcolor{red}{[}his\textcolor{red}{]}$^{\textcolor{red}{1}}$ lips .

\textcolor{lime}{[}The officer\textcolor{lime}{]}$^{\textcolor{lime}{10}}$ led \textcolor{red}{[}Bruce\textcolor{red}{]}$^{\textcolor{red}{1}}$ through \textcolor{olive}{[}the crowd\textcolor{olive}{]}$^{\textcolor{olive}{8}}$ , \textcolor{cyan}{[}the cameras\textcolor{cyan}{]}$^{\textcolor{cyan}{9}}$ capturing every moment .

\textcolor{red}{[}The man in the gray sweater\textcolor{red}{]}$^{\textcolor{red}{12}}$ followed , \textcolor{red}{[}his\textcolor{red}{]}$^{\textcolor{red}{12}}$ presence a comforting anchor .

As they moved through \textcolor{olive}{[}the crowd\textcolor{olive}{]}$^{\textcolor{olive}{8}}$ , \textcolor{red}{[}Bruce\textcolor{red}{]}$^{\textcolor{red}{1}}$ felt a sense of relief wash over \textcolor{red}{[}him\textcolor{red}{]}$^{\textcolor{red}{1}}$ .

\textcolor{orange}{[}The bag , once a symbol of \textcolor{red}{[}his\textcolor{red}{]}$^{\textcolor{red}{1}}$ burden\textcolor{orange}{]}$^{\textcolor{orange}{3}}$ , now felt lighter , as if \textcolor{teal}{[}the weight\textcolor{teal}{]}$^{\textcolor{teal}{4}}$ had been lifted .

In the end , \textcolor{red}{[}Bruce\textcolor{red}{]}$^{\textcolor{red}{1}}$ stood in \textcolor{blue}{[}the dimly lit room\textcolor{blue}{]}$^{\textcolor{blue}{2}}$ , \textcolor{orange}{[}the bag\textcolor{orange}{]}$^{\textcolor{orange}{3}}$ now empty .

\textcolor{brown}{[}The blue lines on \textcolor{purple}{[}the floor\textcolor{purple}{]}$^{\textcolor{purple}{5}}$\textcolor{brown}{]}$^{\textcolor{brown}{6}}$ seemed to glow , a reminder of the journey \textcolor{red}{[}he\textcolor{red}{]}$^{\textcolor{red}{1}}$ had taken .

\textcolor{cyan}{[}The cameras\textcolor{cyan}{]}$^{\textcolor{cyan}{9}}$ were gone , \textcolor{olive}{[}the crowd\textcolor{olive}{]}$^{\textcolor{olive}{8}}$ dispersed , but \textcolor{red}{[}Bruce\textcolor{red}{]}$^{\textcolor{red}{1}}$ felt a sense of peace .

\textcolor{red}{[}He\textcolor{red}{]}$^{\textcolor{red}{1}}$ had faced \textcolor{red}{[}his\textcolor{red}{]}$^{\textcolor{red}{1}}$ fears , and in doing so , had found a new sense of purpose .

        \vspace{.5cm}
        \textbf{\modelname{Gemini 2.0 Flash}}\\
\textcolor{red}{[}A black body bag\textcolor{red}{]}$^{\textcolor{red}{1}}$ was lying on a stretcher .

Someone 's hands were touching \textcolor{blue}{[}the body\textcolor{blue}{]}$^{\textcolor{blue}{2}}$ , perhaps to see if \textcolor{blue}{[}that person\textcolor{blue}{]}$^{\textcolor{blue}{2}}$ was still alive .

Later that day , \textcolor{orange}{[}a coroner\textcolor{orange}{]}$^{\textcolor{orange}{3}}$ was surrounded by a press conference .

\textcolor{orange}{[}The coroner 's\textcolor{orange}{]}$^{\textcolor{orange}{3}}$ actions were probably the focus of the media .

The same scene showed \textcolor{teal}{[}Bruce\textcolor{teal}{]}$^{\textcolor{teal}{4}}$ arguing with \textcolor{purple}{[}a police officer\textcolor{purple}{]}$^{\textcolor{purple}{5}}$ .

\textcolor{teal}{[}He\textcolor{teal}{]}$^{\textcolor{teal}{4}}$ seemed worried , or maybe just angry , but certainly not happy with \textcolor{purple}{[}the officer\textcolor{purple}{]}$^{\textcolor{purple}{5}}$ .

\textcolor{purple}{[}The police officer\textcolor{purple}{]}$^{\textcolor{purple}{5}}$ was not listening , or had \textcolor{purple}{[}his\textcolor{purple}{]}$^{\textcolor{purple}{5}}$ own agenda .

\textcolor{purple}{[}He\textcolor{purple}{]}$^{\textcolor{purple}{5}}$ seemed unwilling to help \textcolor{teal}{[}Bruce\textcolor{teal}{]}$^{\textcolor{teal}{4}}$ , or maybe just was doing \textcolor{purple}{[}his\textcolor{purple}{]}$^{\textcolor{purple}{5}}$ job , blocking access to the crime scene .

Many reporters and cops were present , creating \textcolor{brown}{[}a hectic situation\textcolor{brown}{]}$^{\textcolor{brown}{6}}$ .

\textcolor{brown}{[}The situation\textcolor{brown}{]}$^{\textcolor{brown}{6}}$ seemed to be about \textcolor{blue}{[}the body in \textcolor{red}{[}the black bag\textcolor{red}{]}$^{\textcolor{red}{1}}$\textcolor{blue}{]}$^{\textcolor{blue}{2}}$ and perhaps the circumstances that lead to this tragedy .

\textcolor{teal}{[}Bruce 's\textcolor{teal}{]}$^{\textcolor{teal}{4}}$ body language seemed to indicate that \textcolor{teal}{[}he\textcolor{teal}{]}$^{\textcolor{teal}{4}}$ was not pleased with the current state of things .

        \vspace{.5cm}
        \textbf{\modelname{GPT4o}}\\

\textcolor{red}{[}A mysterious situation\textcolor{red}{]}$^{\textcolor{red}{1}}$ unfolded late one night in the bustling corridors of the city hospital .

A black body bag lay ominously on a stretcher , drawing worried glances .

The coroner , surrounded by flashing cameras and an eager crowd of reporters , quietly spoke to the press , trying to maintain some level of order amidst \textcolor{blue}{[}the chaos\textcolor{blue}{]}$^{\textcolor{blue}{2}}$ .

In the middle of \textcolor{blue}{[}this confusion\textcolor{blue}{]}$^{\textcolor{blue}{2}}$ , \textcolor{orange}{[}Bruce , a concerned onlooker who found himself amidst \textcolor{teal}{[}the growing throng\textcolor{teal}{]}$^{\textcolor{teal}{4}}$\textcolor{orange}{]}$^{\textcolor{orange}{3}}$ , tried to gather information .

\textcolor{orange}{[}His\textcolor{orange}{]}$^{\textcolor{orange}{3}}$ brow furrowed with concern as \textcolor{orange}{[}he\textcolor{orange}{]}$^{\textcolor{orange}{3}}$ spoke directly with \textcolor{purple}{[}a nearby police officer\textcolor{purple}{]}$^{\textcolor{purple}{5}}$ .

Tension buzzed through the air as \textcolor{orange}{[}Bruce\textcolor{orange}{]}$^{\textcolor{orange}{3}}$ pressed for answers .

\textcolor{purple}{[}The officer\textcolor{purple}{]}$^{\textcolor{purple}{5}}$ , surrounded by \textcolor{purple}{[}his\textcolor{purple}{]}$^{\textcolor{purple}{5}}$ colleagues , remained firm but sympathetic .

\textcolor{purple}{[}He\textcolor{purple}{]}$^{\textcolor{purple}{5}}$ communicated with \textcolor{orange}{[}Bruce\textcolor{orange}{]}$^{\textcolor{orange}{3}}$ , explaining the procedural details and what little information could be shared .

\textcolor{purple}{[}He\textcolor{purple}{]}$^{\textcolor{purple}{5}}$ ensured \textcolor{orange}{[}Bruce\textcolor{orange}{]}$^{\textcolor{orange}{3}}$ understood that \textcolor{red}{[}the situation\textcolor{red}{]}$^{\textcolor{red}{1}}$ was under control , even if answers were elusive .

As \textcolor{teal}{[}the crowd\textcolor{teal}{]}$^{\textcolor{teal}{4}}$ continued to swell , \textcolor{brown}{[}reporters\textcolor{brown}{]}$^{\textcolor{brown}{6}}$ pushed closer , eager for any snippets of information \textcolor{brown}{[}they\textcolor{brown}{]}$^{\textcolor{brown}{6}}$ could relay to the world outside .

\textcolor{orange}{[}Bruce\textcolor{orange}{]}$^{\textcolor{orange}{3}}$ stood steadfast amidst the bustle , determined to uncover more .

\textcolor{red}{[}The situation\textcolor{red}{]}$^{\textcolor{red}{1}}$ was intense , but \textcolor{orange}{[}he\textcolor{orange}{]}$^{\textcolor{orange}{3}}$ remained resolute in seeking the truth .

Through \textcolor{blue}{[}the restless chaos\textcolor{blue}{]}$^{\textcolor{blue}{2}}$ , \textcolor{purple}{[}the police officer\textcolor{purple}{]}$^{\textcolor{purple}{5}}$ gestured \textcolor{orange}{[}Bruce\textcolor{orange}{]}$^{\textcolor{orange}{3}}$ to follow \textcolor{purple}{[}him\textcolor{purple}{]}$^{\textcolor{purple}{5}}$ to a quieter corner , hoping to provide clearer insights into \textcolor{red}{[}the situation\textcolor{red}{]}$^{\textcolor{red}{1}}$ .

The night was far from over , and the truth lay hidden within the depths of the shadows .

        \vspace{.5cm}
        \textbf{\modelname{InternVL2.5-78B}}\\

The chaotic scene unfolded with intense fervor , as a diverse group of reporters clamored for information .

The news was breaking , and it was clear that this moment would go down in history .

Amidst the bustling media presence , \textcolor{red}{[}a black police officer , distinguished by his stern expression and authoritative demeanor\textcolor{red}{]}$^{\textcolor{red}{1}}$ , was the focal point of the frenzy—a veritable island of order in the midst of the storm .

Known to many by \textcolor{red}{[}his\textcolor{red}{]}$^{\textcolor{red}{1}}$ name , Bruce , \textcolor{red}{[}he\textcolor{red}{]}$^{\textcolor{red}{1}}$ was well-versed in handling crowd control .

\textcolor{blue}{[}A man in \textcolor{orange}{[}a gray sweater\textcolor{orange}{]}$^{\textcolor{orange}{3}}$\textcolor{blue}{]}$^{\textcolor{blue}{2}}$ approached \textcolor{red}{[}Bruce\textcolor{red}{]}$^{\textcolor{red}{1}}$ with evident urgency , \textcolor{blue}{[}his\textcolor{blue}{]}$^{\textcolor{blue}{2}}$ eyes betraying a mixture of concern and determination .

The exchange between \textcolor{teal}{[}them\textcolor{teal}{]}$^{\textcolor{teal}{4}}$ was brief but tense , encapsulating the gravity of \textcolor{purple}{[}the situation\textcolor{purple}{]}$^{\textcolor{purple}{5}}$ .

With a stern look , \textcolor{red}{[}Bruce\textcolor{red}{]}$^{\textcolor{red}{1}}$ listened intently , a silent sentinel assessing \textcolor{blue}{[}the man 's\textcolor{blue}{]}$^{\textcolor{blue}{2}}$ words with a professional detachment .

As \textcolor{red}{[}Bruce\textcolor{red}{]}$^{\textcolor{red}{1}}$ gave \textcolor{red}{[}his\textcolor{red}{]}$^{\textcolor{red}{1}}$ serious reply , \textcolor{blue}{[}the man in \textcolor{orange}{[}the gray sweater\textcolor{orange}{]}$^{\textcolor{orange}{3}}$\textcolor{blue}{]}$^{\textcolor{blue}{2}}$ seemed to plead earnestly for cooperation .

The urgency was palpable , underscoring the high stakes of \textcolor{teal}{[}their\textcolor{teal}{]}$^{\textcolor{teal}{4}}$ interaction .

The camera lights flickered in the background , casting a stark , almost surreal light on the unfolding drama .

Just then , \textcolor{brown}{[}a man in a leather jacket and cap\textcolor{brown}{]}$^{\textcolor{brown}{6}}$ maneuvered through the crowd , seemingly oblivious to the surrounding commotion .

With a serious expression , \textcolor{brown}{[}he\textcolor{brown}{]}$^{\textcolor{brown}{6}}$ seemed to have just entered from a nearby door , \textcolor{brown}{[}his\textcolor{brown}{]}$^{\textcolor{brown}{6}}$ presence yet another layer of complexity to \textcolor{purple}{[}the situation\textcolor{purple}{]}$^{\textcolor{purple}{5}}$ .

\textcolor{blue}{[}The man in \textcolor{orange}{[}the gray sweater\textcolor{orange}{]}$^{\textcolor{orange}{3}}$\textcolor{blue}{]}$^{\textcolor{blue}{2}}$ looked at \textcolor{red}{[}Bruce\textcolor{red}{]}$^{\textcolor{red}{1}}$ with unflinching resolve as \textcolor{teal}{[}they\textcolor{teal}{]}$^{\textcolor{teal}{4}}$ continued \textcolor{teal}{[}their\textcolor{teal}{]}$^{\textcolor{teal}{4}}$ conversation , the tension between \textcolor{teal}{[}them\textcolor{teal}{]}$^{\textcolor{teal}{4}}$ thickening the electric atmosphere .

\textcolor{magenta}{[}The camera crews , including one with a large ' W ' sign on the back of \textcolor{magenta}{[}their\textcolor{magenta}{]}$^{\textcolor{magenta}{7}}$ jacket\textcolor{magenta}{]}$^{\textcolor{magenta}{7}}$ , were recording every word—a testament to the importance of the discussion .

Amidst all this , \textcolor{olive}{[}another black police officer\textcolor{olive}{]}$^{\textcolor{olive}{8}}$ , \textcolor{olive}{[}his\textcolor{olive}{]}$^{\textcolor{olive}{8}}$ badge gleaming under the harsh lights , managed the chaotic throngs of reporters .

\textcolor{olive}{[}His\textcolor{olive}{]}$^{\textcolor{olive}{8}}$ calm yet vigilant presence served as a counterbalance to the frenzied energy swirling around \textcolor{olive}{[}him\textcolor{olive}{]}$^{\textcolor{olive}{8}}$ .

\textcolor{cyan}{[}The scene\textcolor{cyan}{]}$^{\textcolor{cyan}{9}}$ captured a moment fraught with tension , as \textcolor{teal}{[}\textcolor{blue}{[}the man in \textcolor{orange}{[}the gray sweater\textcolor{orange}{]}$^{\textcolor{orange}{3}}$\textcolor{blue}{]}$^{\textcolor{blue}{2}}$ and \textcolor{red}{[}Officer Bruce\textcolor{red}{]}$^{\textcolor{red}{1}}$\textcolor{teal}{]}$^{\textcolor{teal}{4}}$ navigated a world where the stakes were high and every word counted .

\textcolor{cyan}{[}It\textcolor{cyan}{]}$^{\textcolor{cyan}{9}}$ was a snapshot of history in the making , a testament to the intersection of media , law enforcement , and the relentless pursuit of the truth .

        \vspace{.5cm}
        \textbf{\modelname{Qwen2-VL-72B}}\\

\textcolor{red}{[}Bruce\textcolor{red}{]}$^{\textcolor{red}{1}}$ was lying on the floor , unconscious , after being shot .

\textcolor{red}{[}He\textcolor{red}{]}$^{\textcolor{red}{1}}$ had been involved in a dangerous situation , and things had taken a turn for the worse .

As \textcolor{red}{[}he\textcolor{red}{]}$^{\textcolor{red}{1}}$ lay there , a group of people gathered around \textcolor{red}{[}him\textcolor{red}{]}$^{\textcolor{red}{1}}$ , including \textcolor{blue}{[}a police officer\textcolor{blue}{]}$^{\textcolor{blue}{2}}$ .

\textcolor{blue}{[}The officer\textcolor{blue}{]}$^{\textcolor{blue}{2}}$ was trying to maintain order amidst the chaos .

The scene was chaotic , with cameras flashing and reporters shouting questions .

\textcolor{red}{[}Bruce 's\textcolor{red}{]}$^{\textcolor{red}{1}}$ condition was critical , and \textcolor{blue}{[}the police officer\textcolor{blue}{]}$^{\textcolor{blue}{2}}$ was doing \textcolor{blue}{[}his\textcolor{blue}{]}$^{\textcolor{blue}{2}}$ best to keep everyone at bay .

Despite the commotion , \textcolor{blue}{[}the officer\textcolor{blue}{]}$^{\textcolor{blue}{2}}$ remained calm and collected , ensuring that medical help arrived quickly .

As the situation unfolded , it became clear that \textcolor{red}{[}Bruce\textcolor{red}{]}$^{\textcolor{red}{1}}$ had been targeted .

\textcolor{blue}{[}The police officer\textcolor{blue}{]}$^{\textcolor{blue}{2}}$ questioned witnesses and gathered evidence , determined to find out who was responsible for the attack .

\textcolor{orange}{[}\textcolor{red}{[}Bruce 's\textcolor{red}{]}$^{\textcolor{red}{1}}$ friends and colleagues\textcolor{orange}{]}$^{\textcolor{orange}{3}}$ were shocked and devastated by the news , but \textcolor{orange}{[}they\textcolor{orange}{]}$^{\textcolor{orange}{3}}$ knew that \textcolor{blue}{[}the police officer\textcolor{blue}{]}$^{\textcolor{blue}{2}}$ would do everything in \textcolor{blue}{[}his\textcolor{blue}{]}$^{\textcolor{blue}{2}}$ power to bring \textcolor{teal}{[}the perpetrator\textcolor{teal}{]}$^{\textcolor{teal}{4}}$ to justice .

Despite the danger and uncertainty , \textcolor{blue}{[}the police officer\textcolor{blue}{]}$^{\textcolor{blue}{2}}$ remained steadfast in \textcolor{blue}{[}his\textcolor{blue}{]}$^{\textcolor{blue}{2}}$ duty .

\textcolor{blue}{[}He\textcolor{blue}{]}$^{\textcolor{blue}{2}}$ worked tirelessly to ensure that \textcolor{red}{[}Bruce\textcolor{red}{]}$^{\textcolor{red}{1}}$ received the care \textcolor{red}{[}he\textcolor{red}{]}$^{\textcolor{red}{1}}$ needed and that justice was served .

In the end , thanks to \textcolor{blue}{[}his\textcolor{blue}{]}$^{\textcolor{blue}{2}}$ efforts , \textcolor{teal}{[}the culprit\textcolor{teal}{]}$^{\textcolor{teal}{4}}$ was caught , and \textcolor{red}{[}Bruce\textcolor{red}{]}$^{\textcolor{red}{1}}$ made a full recovery .

\textcolor{blue}{[}The police officer\textcolor{blue}{]}$^{\textcolor{blue}{2}}$ had proven \textcolor{blue}{[}himself\textcolor{blue}{]}$^{\textcolor{blue}{2}}$ to be a true hero , risking \textcolor{blue}{[}his\textcolor{blue}{]}$^{\textcolor{blue}{2}}$ own safety to protect others and uphold the law .

        \vspace{.5cm}
        \textbf{\modelname{Qwen2-VL-7B}}\\

The sequence of images depicts \textcolor{orange}{[}a tense and chaotic scene involving \textcolor{red}{[}a police officer\textcolor{red}{]}$^{\textcolor{red}{1}}$ and \textcolor{blue}{[}a crowd of people\textcolor{blue}{]}$^{\textcolor{blue}{2}}$\textcolor{orange}{]}$^{\textcolor{orange}{3}}$ .

\textcolor{red}{[}The officer\textcolor{red}{]}$^{\textcolor{red}{1}}$ , wearing \textcolor{teal}{[}a police uniform\textcolor{teal}{]}$^{\textcolor{teal}{4}}$ , appears to be in a confrontational situation with \textcolor{blue}{[}a group of individuals , including \textcolor{purple}{[}a man in a gray sweater\textcolor{purple}{]}$^{\textcolor{purple}{5}}$\textcolor{blue}{]}$^{\textcolor{blue}{2}}$ .

\textcolor{blue}{[}The crowd\textcolor{blue}{]}$^{\textcolor{blue}{2}}$ seems to be gathered around \textcolor{red}{[}the officer\textcolor{red}{]}$^{\textcolor{red}{1}}$ , with some people holding cameras and microphones , suggesting a media presence .

\textcolor{brown}{[}The story\textcolor{brown}{]}$^{\textcolor{brown}{6}}$ begins with \textcolor{red}{[}a police officer\textcolor{red}{]}$^{\textcolor{red}{1}}$ standing in the middle of \textcolor{blue}{[}a crowd\textcolor{blue}{]}$^{\textcolor{blue}{2}}$ , surrounded by reporters and photographers .

\textcolor{red}{[}The officer\textcolor{red}{]}$^{\textcolor{red}{1}}$ looks tired and stressed , with dark circles under \textcolor{red}{[}his\textcolor{red}{]}$^{\textcolor{red}{1}}$ eyes .

\textcolor{red}{[}He\textcolor{red}{]}$^{\textcolor{red}{1}}$ is wearing \textcolor{teal}{[}a police uniform\textcolor{teal}{]}$^{\textcolor{teal}{4}}$ and a badge on \textcolor{red}{[}his\textcolor{red}{]}$^{\textcolor{red}{1}}$ chest .

\textcolor{blue}{[}The crowd\textcolor{blue}{]}$^{\textcolor{blue}{2}}$ is , with people asking questions and taking pictures .

As \textcolor{red}{[}the officer\textcolor{red}{]}$^{\textcolor{red}{1}}$ tries to maintain control , \textcolor{purple}{[}a man in \textcolor{magenta}{[}a gray sweater\textcolor{magenta}{]}$^{\textcolor{magenta}{7}}$\textcolor{purple}{]}$^{\textcolor{purple}{5}}$ approaches \textcolor{red}{[}him\textcolor{red}{]}$^{\textcolor{red}{1}}$ .

\textcolor{purple}{[}The man\textcolor{purple}{]}$^{\textcolor{purple}{5}}$ appears to be angry and confrontational , yelling at \textcolor{red}{[}the officer\textcolor{red}{]}$^{\textcolor{red}{1}}$ and \textcolor{blue}{[}the crowd\textcolor{blue}{]}$^{\textcolor{blue}{2}}$ .

\textcolor{red}{[}The officer\textcolor{red}{]}$^{\textcolor{red}{1}}$ remains calm and composed , but \textcolor{olive}{[}the situation\textcolor{olive}{]}$^{\textcolor{olive}{8}}$ escalates quickly .

Suddenly , a camera flash goes off , and \textcolor{red}{[}the officer\textcolor{red}{]}$^{\textcolor{red}{1}}$ turns to look at the source .

\textcolor{red}{[}He\textcolor{red}{]}$^{\textcolor{red}{1}}$ seems to be on edge , as if anticipating trouble .

\textcolor{blue}{[}The crowd\textcolor{blue}{]}$^{\textcolor{blue}{2}}$ continues to press in , and \textcolor{red}{[}the officer 's\textcolor{red}{]}$^{\textcolor{red}{1}}$ body language suggests \textcolor{red}{[}he\textcolor{red}{]}$^{\textcolor{red}{1}}$ is preparing for a physical altercation .

In the midst of \textcolor{cyan}{[}the chaos\textcolor{cyan}{]}$^{\textcolor{cyan}{9}}$ , \textcolor{lime}{[}a police officer in \textcolor{violet}{[}a hat and uniform\textcolor{violet}{]}$^{\textcolor{violet}{11}}$\textcolor{lime}{]}$^{\textcolor{lime}{10}}$ steps forward , trying to calm \textcolor{olive}{[}the situation\textcolor{olive}{]}$^{\textcolor{olive}{8}}$ .

\textcolor{lime}{[}He\textcolor{lime}{]}$^{\textcolor{lime}{10}}$ speaks firmly to \textcolor{blue}{[}the crowd\textcolor{blue}{]}$^{\textcolor{blue}{2}}$ , but the tension remains high .

\textcolor{purple}{[}The officer in \textcolor{magenta}{[}the gray sweater\textcolor{magenta}{]}$^{\textcolor{magenta}{7}}$\textcolor{purple}{]}$^{\textcolor{purple}{5}}$ continues to argue , and \textcolor{blue}{[}the crowd\textcolor{blue}{]}$^{\textcolor{blue}{2}}$ seems to be divided , with some people supporting \textcolor{red}{[}the officer\textcolor{red}{]}$^{\textcolor{red}{1}}$ and others siding with \textcolor{purple}{[}the man in \textcolor{magenta}{[}the gray sweater\textcolor{magenta}{]}$^{\textcolor{magenta}{7}}$\textcolor{purple}{]}$^{\textcolor{purple}{5}}$ .

As \textcolor{orange}{[}the scene\textcolor{orange}{]}$^{\textcolor{orange}{3}}$ reaches \textcolor{orange}{[}its\textcolor{orange}{]}$^{\textcolor{orange}{3}}$ climax , \textcolor{lime}{[}the officer in \textcolor{violet}{[}the hat and uniform\textcolor{violet}{]}$^{\textcolor{violet}{11}}$\textcolor{lime}{]}$^{\textcolor{lime}{10}}$ tries to intervene , but \textcolor{olive}{[}the situation\textcolor{olive}{]}$^{\textcolor{olive}{8}}$ quickly degrades into a physical altercation .

\textcolor{purple}{[}The officer in \textcolor{magenta}{[}the gray sweater\textcolor{magenta}{]}$^{\textcolor{magenta}{7}}$\textcolor{purple}{]}$^{\textcolor{purple}{5}}$ appears to be physically attacking \textcolor{red}{[}the police officer\textcolor{red}{]}$^{\textcolor{red}{1}}$ , and \textcolor{blue}{[}the crowd\textcolor{blue}{]}$^{\textcolor{blue}{2}}$ becomes increasingly hostile .

\textcolor{brown}{[}The story\textcolor{brown}{]}$^{\textcolor{brown}{6}}$ ends with \textcolor{red}{[}the police officer\textcolor{red}{]}$^{\textcolor{red}{1}}$ being overwhelmed by \textcolor{blue}{[}the crowd\textcolor{blue}{]}$^{\textcolor{blue}{2}}$ , and \textcolor{olive}{[}the situation\textcolor{olive}{]}$^{\textcolor{olive}{8}}$ spiraling out of control .

\textcolor{lime}{[}The officer in \textcolor{violet}{[}the hat and uniform\textcolor{violet}{]}$^{\textcolor{violet}{11}}$\textcolor{lime}{]}$^{\textcolor{lime}{10}}$ tries to help , but \textcolor{cyan}{[}the chaos\textcolor{cyan}{]}$^{\textcolor{cyan}{9}}$ is too much to handle .

\textcolor{orange}{[}The scene\textcolor{orange}{]}$^{\textcolor{orange}{3}}$ is a vivid portrayal of a tense and dangerous situation , with \textcolor{red}{[}the police officer\textcolor{red}{]}$^{\textcolor{red}{1}}$ struggling to maintain order in the face of \textcolor{blue}{[}a hostile crowd\textcolor{blue}{]}$^{\textcolor{blue}{2}}$ .

\end{tcolorbox}

\twocolumn

\end{document}